%% file: main-aalto.tex
\documentclass[sigconf]{acmart}

\setcopyright{none} 
\renewcommand\footnotetextcopyrightpermission[1]{} 

\usepackage{xspace}
\usepackage{tabularx}
\usepackage{hyperref}
\usepackage{amsmath}
\usepackage{multirow}
\usepackage{subfig}
\usepackage{algorithm}
\usepackage{algorithmicx}
\usepackage{algpseudocode}
\usepackage{todonotes}
\usepackage{eurosym}

\input{macros}
\usepackage{enumitem}

\acmConference[ACSAC '19]{2019 Annual Computer Security Applications Conference}{December 9--13, 2019}{San Juan, PR, USA}
\acmDOI{10.1145/3359789.3359810}
\acmISBN{978-1-4503-7628-0/19/12}

\acmBadgeR[https://www.acsac.org/2019/]{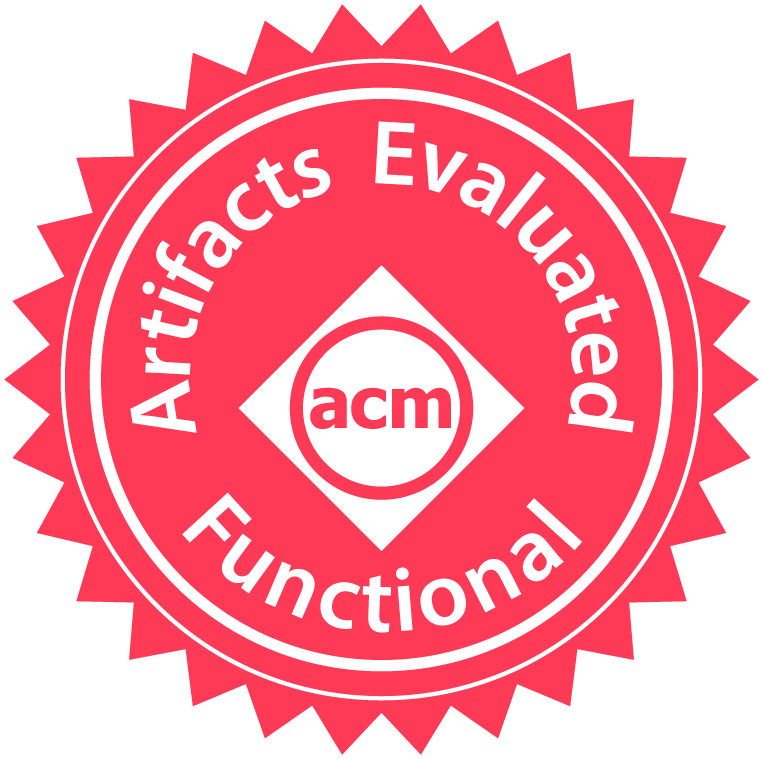}

\begin{document}
\title{Detecting organized eCommerce fraud \\ using scalable categorical clustering}

\author{Samuel Marchal}
\orcid{0000-0002-8522-2707}
\affiliation{%
  \institution{Aalto University}
}
\email{samuel.marchal@aalto.fi}

\author{Sebastian Szyller}
\affiliation{%
  \institution{Aalto University}
}
\email{sebastian.szyller@aalto.fi}

\begin{abstract}
\input{abstract}
\end{abstract}



\maketitle

\pagestyle{plain} 

\input{intro}
\input{pb_statement}
\input{design}
\input{weight}
\input{exp_setup}
\input{exp_agglo}
\input{exp_recursive}
\input{exp_real_world}
\input{related-work}
\input{conclusion}

\section*{Acknowledgments}
This research was funded by a research donation from Zalando Payments GmbH. It is supported by the Academy of Finland through the SELIoT Project (Grant 309994). We thank Zalando employees and N. Asokan for interesting discussions and valuable feedback.

\bibliographystyle{ACM-Reference-Format}
\bibliography{main-aalto}

\appendix
\input{app-clustering}
\input{app-dataset}
\input{app-weighting}
\input{app-samplClust}

\end{document}

%% file: macros.tex
\newcommand{\ourname}{\textproc{RecAgglo}\xspace}

\newcommand{\order}{order\xspace}
\newcommand{\orders}{orders\xspace}
\newcommand{\Order}{Order\xspace}

\newcommand{\fraud}{fraud\xspace}
\newcommand{\Fraud}{Fraud\xspace}

\newcommand{\frtest}{\textsf{TestF-15K}\xspace}
\newcommand{\frtrain}{\textsf{TrainF-15K}\xspace}
\newcommand{\gesmall}{\textsf{TrainG-30K}\xspace}
\newcommand{\gelarge}{\textsf{TrainG-100K}\xspace}

\newcommand{\gebg}{\textsf{DE-bg}\xspace}
\newcommand{\gereal}{\textsf{DE-real}\xspace}
\newcommand{\frbg}{\textsf{FR-bg}\xspace}
\newcommand{\frreal}{\textsf{FR-real}\xspace}
\newcommand{\hrbg}{\textsf{CH-bg}\xspace}
\newcommand{\hrreal}{\textsf{CH-real}\xspace}
\newcommand{\nlbg}{\textsf{NL-bg}\xspace}
\newcommand{\nlreal}{\textsf{NL-real}\xspace}
\newcommand{\bebg}{\textsf{BE-bg}\xspace}
\newcommand{\bereal}{\textsf{BE-real}\xspace}

\newcommand*\mean[1]{\overline{#1}}
\newcommand{\mathrange}[1]{\mathit{[{#1}]}}

\usepackage{xcolor}

\newcommand{\oldtext}[1]{}

\newcommand*\circled[1]{\tikz[baseline=(char.base)]{
            \node[shape=circle,fill,inner sep=1pt] (char) {\textcolor{white}{#1}};}}



%% file: abstract.tex
Online retail, eCommerce, frequently falls victim to fraud conducted by malicious customers (fraudsters) who obtain goods or services through deception.
Fraud coordinated by groups of professional fraudsters that place several fraudulent orders to maximize their gain is referred to as \textit{organized fraud}.
Existing approaches to fraud detection typically analyze orders in isolation and they are not effective at identifying groups of fraudulent orders linked to organized fraud.
These also wrongly identify many legitimate \order{s} as fraud, which hinders their usage for automated fraud cancellation.
We introduce a novel solution to detect organized fraud by analyzing orders in bulk.
Our approach is based on clustering and aims to group together fraudulent orders placed by the same group of fraudsters.
It selectively uses two existing techniques, \textit{agglomerative clustering} and \textit{sampling} to \textit{recursively} group orders into small clusters in a reasonable amount of time.
We assess our clustering technique on real-world orders placed on the Zalando website, the largest online apparel retailer in Europe\footnote{Disclaimer:
The views and opinions expressed in this article are those of the authors and do not
necessarily reflect the official position of Zalando Payments GmbH.}. Our clustering processes 100,000s of orders in a few hours and groups 35-45\% of fraudulent \order{s} together. We propose a simple technique built on top of our clustering that detects 26.2\% of fraud while raising false alarms for only 0.1\% of legitimate orders.


%% file: intro.tex
\section{Introduction}

Online retail, also known as eCommerce, represents an important share of the retail business.
About 17.5\% of all sales made in the United States consists of eCommerce transactions, which accounts for several trillions of dollars every year~\cite{online-sales:2019}.
The expansion of online retail is driven by its two main characteristics: 24/7 accessibility and scalability to a potentially unlimited number of customers.
However, these features also increase the exposure to \textit{frauds} in which malicious customers, \textit{fraudsters}, obtain physical goods or services through deception.
It is estimated that 3 to 5\% of online \order{s} constitute fraud, which accounts for over \$50B in value every year~\cite{GFI:2017}.
Fraud represents a direct monetary loss that can significantly reduce the business valuation of online retailers~\cite{zalando-fraud:2015:2} and it must be mitigated.

Cancellation of fraudulent \order{s} in a timely manner  prevents this monetary damage.
To be effective, \fraud cancellation requires reliable means of detection.
Cancelling legitimate \order{s} decreases customer loyalty, degrades brand image/reputation and causes a shortfall in revenue estimated to over \$100B every year~\cite{false-decilne:2017}.
Because of this reliability requirement, current approaches to canceling \fraud rely on \textit{screening} which is a manual process done by human analysts~\cite{screening:2014}.
Thus, screening is a costly process applied to a limited number of \order{s} and which can prevent only a limited amount of fraud. 
Screening can be facilitated using automated analysis techniques that produce additional fraud indicators~\cite{mao2018adaptive}.
Nevertheless, these techniques are not accurate enough to provide standalone and automated fraud cancellation~\cite{screening:2014}.

The type of fraud that has been rising is \textit{organized fraud}~\cite{levi2008organized}.
95 professional fraudsters performing organized fraud were arrested in 2018 for committing fraud exceeding \euro8M in value~\cite{organized-fraud:2018}.
In organized fraud, a small group of fraudsters coordinate \textit{fraud campaigns} against a chosen online retailer. Fraud campaigns span a limited period of time during which several fraudulent \order{s} are placed for goods delivered in a limited geographical area.
The online retailer, Zalando, lost \euro18.5M to organized fraud in one quarter of 2015~\cite{zalando-fraud:2015:2}.
Since then, Zalando invested systematically into their fraud detection systems, which
reduced fraud to very low numbers. Nevertheless, throughout the market, many current automated techniques for fraud detection analyze orders in isolation~\cite{sorin2012fraudsurvey, bolton2002fraudreview, carcillo2017spark, gomez2018neuralnetsfraud, levi2008organized}.
However, detecting organized fraud profits from a global view of all orders placed in a given period of time.

\textbf{Goal and contributions.}
We want to design a solution to detect organized eCommerce fraud.
We propose analyzing \order{s} in bulk rather than in isolation, to identify similarities among fraudulent \order{s} that belong to the same fraud campaign.
Similar fraudulent orders can be grouped together by applying \textit{clustering} on relevant attributes.
\Order{s} in the same fraud campaign have common characteristics highlighted by identical \textit{categorical attributes}, e.g., delivery address, customer name, payment method, etc.
Consequently, we propose applying unsupervised \textit{categorical clustering} on these attributes to identify organized eCommerce fraud.

We introduce a novel approach for hierarchical categorical clustering: \textit{recursive agglomerative clustering}.
This approach is specifically designed to group fraudulent \order{s} placed on online retail stores.
It combines the benefits of (1) \textit{agglomerative clustering} to generate small clusters each potentially representing a fraud campaign and (2) \textit{sampling} to process a large number of \order{s} in a reasonable amount of time.
Clusters obtained using our clustering approach have two applications: \ref{app:screening} prioritizing \order{s} that must be analyzed through screening and \ref{app:automated} automatically canceling frauds by deciding that all \orders in a cluster are fraudulent if at least one \order in the cluster is fraudulent (e.g., older known fraud).

We assess the real-world effectiveness of our approach using 6M \order{s} placed on the Zalando website~\cite{zalando2019landing}, the largest online apparel retailer in Europe.
We claim the following contributions:
\begin{itemize}
		\item a novel clustering technique for categorical data (Sect.~\ref{sec:recagglo}). It is a recursive clustering approach combining agglomerative clustering and sampling to generate a large number of clusters from medium-size datasets (100,000s of samples).
		\item two strategies for weighting categorical attributes (Sect.~\ref{sec:weight}), which facilitate the selection of optimal hyperparameters for categorical clustering (Sect.~\ref{sec:weight_perf}). 
	\item the evaluation of our clustering technique showing its effectiveness at grouping fraudulent \order{s} (Sect.~\ref{sec:eval_recursive}). It generates clusters mixing a small number of fraudulent and legitimate \order{s} (0.8\%) while grouping a large portion of fraudulent \orders (42.1\%) together. Its computation time is much lower than existing clustering techniques and is able to process, e.g., 15,000 \order{s} in 3 minutes.
	\item the demonstration that generated clusters can be used to automatically detect 26.2\% of real-world fraud perpetrated against Zalando, while causing only 0.1\% false alarms for legitimate \order{s} incorrectly identified as fraud (Sect.~\ref{sec:real-world-eval}).
\end{itemize}

%% file: pb_statement.tex
\section{Detecting eCommerce fraud}

We focus on the detection of fraudulent \orders committed against online retailers, which we simply name \textit{frauds} from now on.
In this paper, we tackle the specific use case of detecting \fraud perpetrated against Zalando~\cite{zalando2019landing}.
Zalando is an online apparel retailer operating in 17 markets, having 28 million active customers and generating over \euro5B in revenue yearly~\cite{zalando2019}.
We believe this use case is representative of many large online retailers.
We focus on detecting \fraud where fraudsters obtain physical goods through delivery with no intent of paying for them.
The \order can either be in payment default or paid with illegally acquired means of payment.
In both cases, the retailer suffers monetary losses. 

\subsection{Fraud detection and cancellation}

\begin{figure}[th]
                \centering
                \includegraphics[width=\columnwidth]{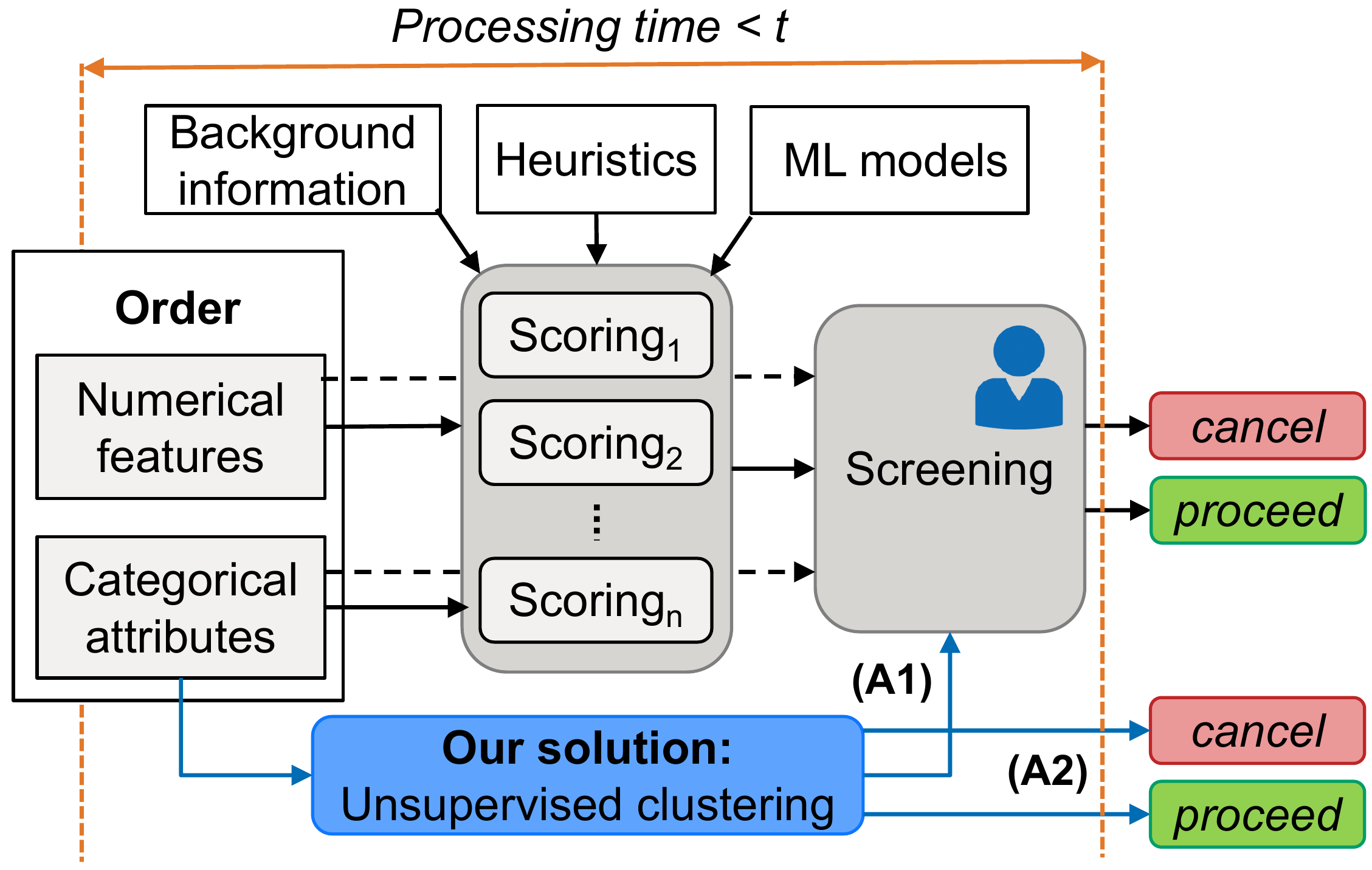}
                \caption{Fraud detection pipeline. Final cancellation is decided by human analysts during screening. Our solution  supports screening (\ref{app:screening}) and automatically cancel frauds (\ref{app:automated}).}
                \label{fig:pipeline}
\end{figure}

A typical\footnote{This pipeline is chosen for the sake of generalizability. The particular fraud detection setup at Zalando is not taken into account in this paper and it does not perfectly match this typical pipeline.} fraud detection pipeline~\cite{screening:2014,chmielewski2009mortgage,8038008} is depicted in Fig.~\ref{fig:pipeline}.
An \order is represented by a set of numerical features and categorical attributes.
This information is automatically validated to confirm the order which serves as a preliminary step for the fraud detection process.
Features and attributes representing an \order are fed to several \textit{scoring functions} that automatically produce fraud indicators.
These functions can use additional background information (e.g., from customer history) and they typically rely on human defined heuristics and supervised machine learning (ML) models~\cite{sorin2012fraudsurvey, bolton2002fraudreview, niu2019comparison}.
Fraud indicators and raw \order information are provided to a screening process that decides if the \order is legitimate and should proceed or if it is a fraud and it should be canceled.
Cancellations are usually performed based on a combination of machine learning systems and
human expert knowledge.

Fraud detection is a time constrained process that must happen after an order is placed and before it is processed for shipping.
This typically gives only a few hours to detect fraud and only a small
fraction of orders can be inspected by human experts. On the other hand, many automated scores
are computed on each order independently, which can decrease their efficiency at detecting organized \fraud.

\subsection{Preventing organized fraud}
\label{sec:organized_fraud}

Fraud can be either isolated events occasionally performed by individuals or organized by criminal groups of professional fraudsters~\cite{levi2008organized}.
Organized fraud relies on coordinated events, \textit{fraud campaigns}, that target a specific online retailer.
During a fraud campaign, several \order{s} are placed over a limited period of time (e.g., one month), by a small group of fraudsters having several electronic identities each.
Most \order{s} are fraudulent and all \order{s} are delivered in a restricted geographical area (e.g., the same city) where the criminal group operates.
Also, fraud campaign typically uses payment methods that are known to be vulnerable to fraud~\cite{zalando-fraud:2015:2}.
We propose to prevent organized fraud by identifying fraud campaigns.

Following that \fraud belonging to a same campaign has similar characteristics, we propose to group similar \order{s} together in order to identify fraud campaigns.
In this study, we restrict ourselves to certain categorical attributes, i.e., delivery address, customer name, payment method, etc.
Numerical features have already been extensively used for fraud detection~\cite{Brause1999neuraldatamining,carcillo2017spark,thiprungsri2011clusteranalysis} and we want to investigate the capabilities of categorical attributes in their own right.
Since we do not have a priori knowledge about \order{s} that belong to a fraud campaign, we propose to take an unsupervised clustering approach to group similar \order{s} and apply it to categorical attributes of \order{s}.
Ideally it would generate one cluster per fraud campaign, containing all frauds of this campaign but no legitimate \order.
Also, most legitimate \order{s} should have low similarity between each other and thus, be less likely to be grouped into clusters.

Figure~\ref{fig:pipeline} depicts the deployment of our clustering approach in the fraud detection pipeline. It has two applications.

\begin{enumerate}[label=\textbf{A\arabic*}]
  \item \label{app:screening} \textbf{Prioritizing screening.} Clustered \order{s} can be screened with high priority to detect a large number of \fraud{s} with a minimal effort.
  We expect frauds to be clustered at a significantly higher rate than legitimate \order{s}. Comparing \order{s} in the same cluster provides human analysts with new information that may facilitate the cancellation.
  \item \label{app:automated}\textbf{Automated fraud cancellation.} Fraud detection can be applied to clusters of \order{s} rather than to individual \order{s}.
  Several \order{s} provide aggregated information that may depict fraudulent behavior more reliably than isolated \order{s}.
  An automated process can decide if the whole cluster is fraudulent and cancel the \order{s}.
\end{enumerate}

While a significant share of frauds is organized and may be associated with a fraud campaign, frauds can also be isolated events.
We focus on detecting organized fraud only and our approach is not designed to detect isolated fraud cases.
Our approach is complementary to the extensive prior work addressing the detection of fraud in isolation~\cite{gomez2018neuralnetsfraud,aleskervo1997cardwatch,Brause1999neuraldatamining, maes2002bayesandneural}.

\subsection{Attributes representing \order{s}}
\label{sec:attributes}

In the following, we represent each \order by 37 categorical attributes, which have discrete values with no intrinsic ordering.
These attributes were selected for the sake of generalizability.
They belong to 5 categories of information that is generally provided by a customer placing an \order on any online retailer.
\begin{itemize}
	\item Customer $A_{cust}$ (9 attributes): related to the electronic identity of the customer, e.g., email address, IP address, etc.
	\item Delivery $A_{del}$ (3 attributes): related to the means used for \order delivery, e.g, pickup point, delivery type, etc.
	\item Shipping $A_{ship}$ (7 attributes): related to identity and location (address) of the person receiving the \order.
	\item Payment $A_{pay}$ (11 attributes): related to payment method, e.g., bank transfer, credit card suffix, etc.
	\item Billing $A_{bill}$ (7 attributes): related to identity and location (address) of the person paying the \order.
\end{itemize}

Many of our attributes contain Personally Identifiable Information (PII) which were anonymized prior to perform any data analysis. The clustering approach we introduce and the experimental results we obtain use these anonymized attribute values.

\subsection{Challenges in clustering fraud campaigns}
\label{sec:challenges}

Clustering \order{s} that belong to fraud campaigns requires to address several challenges related to (a) categorical clustering, (b) fraud detection and (c) the application domain of online retail.

\begin{enumerate}[label=\textbf{C\arabic*}]
  \item \label{chal:cardinality} \textbf{Imbalanced attribute cardinality.} Categorical attributes representing \orders take a different number of values (cardinality), from two values to millions. High cardinality prevents the numerical encoding of attributes. Imbalance makes it difficult to quantify the similarity between two \orders.
  \item \label{chal:imbalance} \textbf{Imbalanced classes.} The ratio of fraudulent to legitimate \orders is highly imbalanced, typically around 1/50~\cite{GFI:2017}. Probability of clustering legitimate \orders is much higher than that of clustering frauds, which is undesirable. 
  \item \label{chal:groundtruth} \textbf{No ground truth for fraud campaign.} There is no information which fraud corresponds to which fraud campaign. Only ground truth for individual orders is available.
  \item \label{chal:scale} \textbf{Scale of the data.} Large online retailers receive 100,000s of \orders per day. Zalando receives 300,000 \orders on average every day~\cite{zalando2019}. Most existing categorical clustering methods~\cite{ganti1999cactus,guha2000rock,zaki2005clicks} have a high complexity and they cannot process data of such a scale in a reasonable amount of time.

\end{enumerate}

\subsection{Requirements}

We define the following requirements for a clustering approach to detect fraud campaigns:

\begin{enumerate}[label=\textbf{R\arabic*}]

  \item \label{goal:cl_size} \textbf{Generate small clusters}. There are many more legitimate \order{s} than frauds (\ref{chal:imbalance}). Also, a fraud campaign typically contains a low number (e.g., 10s-100s) of \order{s}. In order to group frauds, clustering must generate a large number of small clusters, each potentially corresponding to a single fraud campaign.

  \item \label{goal:impurity} \textbf{Minimize cluster impurity}. The cluster impurity must be low. Generated clusters must be composed either only of legitimate \orders or only of frauds.

  \item \label{goal:fraud} \textbf{Maximize clustered fraud}. Frauds isolated in singletons (clusters with one component) are not linked to any fraud campaign and they cannot be detected by our method. We must maximize the rate of detected fraud.

  \item \label{goal:time} \textbf{Minimize execution time}. Online retailers receive 100,000s of \orders per day. Our approach must be able to process such amount of data (\ref{chal:scale}) in a reasonable amount of time allowing for cancellation (e.g., a few hours).

\end{enumerate}

Regarding \ref{goal:cl_size}, we do not have ground truth for fraud campaigns (\ref{chal:groundtruth}) and we cannot evaluate the ``\textit{goodness}'' of our clusters with respect to grouping frauds from the same campaign.
We ensure clustering \textit{goodness} by requiring a minimum similarity between elements belonging to the same cluster, which is typical in clustering~\cite{maimon2005data}.
This guarantee is provided by enforcing a maximum distance between elements that compose the same cluster.
Keeping this distance below a threshold is our criterion for cluster goodness.



%% file: design.tex
\section{Recursive Agglomerative Clustering}
\label{sec:design}

We introduce \textit{Recursive Agglomerative Clustering} (\textproc{RecAgglo}) a novel approach for categorical clustering.
It combines the benefits of two existing techniques~\cite{maimon2005data}: \textit{agglomerative clustering}, which is able to generate small clusters (\ref{goal:cl_size}) and \textit{sampling}, which reduces the time complexity of clustering methods (\ref{goal:time}).
These two techniques are selectively applied to recursively divide a large set of samples into small clusters which eventually meet our goodness criterion.
The code for the \textproc{RecAgglo} algorithm is publicly available~\cite{recagllo:2019}.

\subsection{Agglomerative clustering and sampling}
\label{sec:agglo_sampling}

\textbf{Agglomerative clustering} is a bottom up hierarchical clustering approach.
Each element is initially placed into a singleton cluster.
Pairs of clusters having the smallest distance to each other are then sequentially merged into larger clusters until all elements are in a single cluster.
The distance between two clusters is defined using a \textit{linkage} method. For instance, \textit{single linkage} uses the minimum distance between any two points in each cluster.
The algorithm produces a dendrogram which represents consecutive merges.
Using the dendrogram, a desired clustering (set of clusters) can be chosen using different criteria, e.g.,  \textit{'distance'}: the maximum distance $d_{max}$ between elements in a cluster,  \textit{'maxclust'}: the maximum number of clusters $c_{max}$ to generate.
In contrast to many clustering techniques~\cite{daniel2002coolcat,huang1998extensions}, agglomerative clustering does not generate a predefined number of clusters.
It generates clusters by grouping the most similar singletons first and it can create many small clusters which meet our goodness criterion as defined by the \textit{'distance'} $d_{max}$. Elements that cannot be assigned to any cluster while meeting this criterion remain isolated in singletons.
Agglomerative clustering requires computing pairwise distances between elements ($O(n^2)$ complexity) and does not scale to large datasets.
The \textproc{AggloClust} algorithm is presented in App.~\ref{app:agglo_clust}. It takes a set $c$ of elements to cluster and a distance $d_{max}$ as inputs.

\noindent\textbf{The sampling algorithm} is applied on top of existing clustering techniques.
It selects a random sample of reference elements from a set of $n$ elements. These reference elements are clustered and the remaining ones (not sampled) are assigned to the initially formed clusters. This reduces the number of distance computations between elements from $n \times n$ to the sample size $\times n$. If the sample size is in the order of $O(log(n))$, the complexity of the base clustering algorithm is reduced by the same factor.

We use sampling to reduce the complexity of agglomerative clustering to $O(n \times log(n))$ by using a sample size in the order of $O(log(n))$.
Our algorithm for agglomerative clustering with sampling \textproc{SampleClust} is detailed in App.~\ref{app:sample_clust}.
It takes 3 inputs: a set $c$ of elements to cluster, $\rho_s$ a multiplying factor to $\sqrt{|c|}$ defining the sample size and $\rho_{mc}$ the maximum number of clusters to generate (using the \textit{'maxclust'} criterion for cluster generation).
$\rho_s$ and $\rho_{mc}$ are parameters to be defined according to the desired computation time.
Sampling deprives agglomerative clustering of its ability to generate many small clusters since the number of clusters is bounded by the sample size. Also, clusters generated using sampling do not meet any goodness criterion defined by the maximum \textit{'distance'} $d_{max}$.


\subsection{Our \textproc{RecAgglo} algorithm}
\label{sec:recagglo}

We introduce a scalable approach to generating clusters that meet our goodness criterion, i.e., the distance between elements in the same cluster is lower than $d_{max}$.
Our solution \circled{1} recursively divides large clusters into smaller ones using \textproc{SampleClust}.
When clusters are small enough, \circled{2} it runs \textproc{AggloClust} to generate a clustering in which each cluster meets the \textit{'distance'} criterion $d_{max}$. All resulting clusters are recursively aggregated to form the final clustering $C_{res}$ composed of clusters that meet our goodness criterion.

\begin{algorithm}[ht]
\caption{Recursive agglomerative clustering}
\label{algo:recursive}
\begin{algorithmic}[1]

	\State{
		Let $C = {c_1, \ldots, c_n}$ denote a clustering of $|C| = n$ clusters, \\
		$c = {v_1, \ldots, v_m}$, a cluster of $|c| = m$ elements $v$, \\
		$\delta_a$, the threshold for using \textproc{AggloClust},\\
		$d_{max}$, the maximum distance for cluster fusion,\\
		$\rho_s$, the multiplying factor for the sample size,\\
		$\rho_{mc}$, the dividing factor for maxclust number.
		} \\

\Function{RecAgglo}{$C, \delta_a, d_{max}, \rho_s, \rho_{mc}$}
	\State{$C_{res} \gets \emptyset$}
	\State{$remain \gets \emptyset$}
	\State{$max_{s} \gets 4 \times \delta_a$} \\

	\For{$c : c \in C$} \hfill \Comment{Loop to split existing clusters}
		\If {$|c| > \delta_a$} 	 \hfill \Comment{Cluster sampling}
			\State{$C_{s} \gets$ \textproc{SampleClust}($c, \rho_s, \rho_{mc}$)}
			 \If {$|C_{s}| > 1$} \hfill \Comment{Recursive clustering}
			 	\State{$C_{loop} \gets$ \textproc{RecAgglo}($C_{s}, \delta_a, d_{max}, \rho_s, \rho_{mc}$)}
			 \ElsIf{$\rho_{mc} > 1.01$} \hfill \Comment{Recursive clustering alt.}
			 	\State{$\rho_{mc} \gets 1.01$} \hfill \Comment{Set higher maxclust}
			 	\State{$C_{loop} \gets$ \textproc{RecAgglo}($C_{s}, \delta_a, d_{max}, \rho_s, \rho_{mc}$)}
			 \ElsIf{$|c| < max_{s}$} \hfill \Comment{Fall back agglomerative}
			 	\State{$C_{loop} \gets$ \textproc{AggloClust}($c, d_{max}$)}
			 \Else \hfill \Comment{No split possible / to re-cluster}
			 	\State{$remain \gets remain \cup c$}
			 \EndIf
		\ElsIf{$|c| > 1$}  \hfill \Comment{Agglomerative clustering}
			\State{$C_{loop} \gets$ \textproc{AggloClust}($c, d_{max}$)}
		\Else \hfill \Comment{Elements to re-cluster}
			\State{$remain \gets remain \cup c$}
		\EndIf
		\State{$C_{res} \gets C_{res} \cup C_{loop}$} \hfill  \Comment{Add new clusters to result}
	\EndFor \\

	\State{\# Clustering non-clustered elements ($remain$)}
	\If {$|remain| > \delta_a$} 	 \hfill \Comment{Cluster sampling}
		\State{$C_{sample} \gets$ \textproc{SampleClust}($remain, \rho_s, \rho_{mc}$)}
		\If {$|C_{sample}| > 1$} \hfill \Comment{Recursive clustering}
			 \State{$C_{end} \gets$ \textproc{RecAgglo}($C_{sample}, \delta_a, d_{max}, \rho_s, \rho_{mc}$}
		\Else \hfill \Comment{Elements are singletons}
			\State{$C_{end} \gets  \lbrace remain \rbrace$} 
		\EndIf
	\ElsIf{$|remain| > 1$}  \hfill \Comment{Agglomerative clustering}
		\State{$C_{end} \gets$ \textproc{AggloClust}($remain, d_{max}$)}
	\Else \hfill \Comment{Elements are singletons}
		\State{$C_{end} \gets  \lbrace remain \rbrace$} 
	\EndIf
	\State{$C_{res} \gets C_{res} \cup C_{end}$}

	\State{return $C_{res}$}
\EndFunction

\end{algorithmic}
\end{algorithm}

Our approach \textproc{RecAgglo} is defined in Algorithm~\ref{algo:recursive}.
It takes as inputs an initial clustering $C$ and a set of parameters: $\delta_a$ (for \textproc{RecAgglo}), $d_{max}$ (for \textproc{AggloClust}), $\rho_s$ and $\rho_{mc}$ (for \textproc{SampleClust}).
$\delta_a$ is a parameter defined according to computation time restrictions.
\textproc{RecAgglo} loops over clusters $c \in C$ to split them into smaller clusters.

If the size of $c$ is larger than a threshold $\delta_a$, $c$ is split using \textproc{SampleClust}.
We use the \textit{'maxclust'} criteria to generate clusters small enough to be eventually processed using \textproc{AggloClust}.
The resulting clustering $C_s$ does not meet our goodness criterion yet and it is re-processed using \textproc{RecAgglo}.
We observed that \textproc{SampleClust} may not be able to split an input set $c$ given a specific 'maxclust' factor $\rho_{mc}$.
We address this in two ways. Firstly, we re-try \textproc{SampleClust} with a lower value $\rho_{mc} = 1.01$ providing the ability to generate more clusters.
Secondly, we fall back to plain agglomerative clustering given that the cluster size is still reasonably low ($4 \times \delta_a$).
Both these measures were determined empirically.
Alternatively, $\rho_{mc}$ could be progressively decreased or the multiplying factor of $\delta_a$ can be changed.
If both measures fail to split $c$ in clusters, elements in $c$ are added to the set $remain$ for further processing.
Alternatively, we obtain a resulting clustering $C_{loop}$ from \textproc{RecAgglo} or \textproc{AggloClust} that meets our goodness criterion.

If the size of $c$ is lower than $\delta_a$ but larger than $1$, $c$ is split using \textproc{AggloClust} with parameter $d_{max}$.
If $c$ is a singleton, it is added to the $remain$ set for later processing.
During each iteration, we add the new clustering $C_{loop}$ to the final clustering $C_{res}$ or we complement the $remain$ set of elements for reprocessing.

The $remain$ set contains clustered elements resulting from \textproc{SampleClust} and singletons - obtained due to lack of sufficiently similar elements in the drawn sample.
Thus, we try to re-cluster these remaining elements, following the same steps as previously.
We use \textproc{SampleClust} (if $|remain| > \delta_a$), \textproc{AggloClust} (if $\delta_a \geq |remain| > 1$) or keep a singleton (if $|remain| = 1$).
If \textproc{SampleClust} is successful at splitting $remain$, we recursively run \textproc{RecAgglo} on the resulting clustering. However, in contrast to the previous process, we do not apply alternative measures if \textproc{SampleClust} fails and just keep all elements as singletons.

The resulting clustering $C_{end}$ is added to $C_{res}$ which is our final clustering where all clusters meet our goodness criterion. It is worth noting that many of these clusters may be singletons.

\subsection{\textproc{RecAgglo} properties}
\label{sec:algo_properties}

\textbf{Achieving cluster goodness:} \textproc{RecAgglo} uses agglomerative clustering  to generate the final clustering $C_{res}$.
Consequently, any cluster in $C_{res}$ of two or more elements meets our goodness criterion defined by the maximum distance $d_{max}$. These clusters are smaller than $\delta_a$ and meet \ref{goal:cl_size} for a sensible choice of $\delta_a$.

\noindent\textbf{Computational complexity}: The complexity of \textproc{RecAgglo} depends on its recursive nature and \textproc{SampleClust} complexity.
\textproc{AggloClust} runs on sets of size with the static upper bound $\delta_a$. Its running time is bounded by a constant.
The maximum complexity of \textproc{SampleClust} during the initial run is $O(n \times log(n))$ and it decreases during subsequent recursions.
In the worst-case scenario, we require at most $n$ recursions to obtain the final clustering. This makes the worst-case complexity of $O((n \times log(n))^{n})$ for \textproc{RecAgglo}.
This theoretical complexity is completely untractable and \textproc{RecAgglo} cannot scale to large datasets in theory.
However, we show in Sect.~\ref{sec:perf_comparison} that its actual complexity is sub-quadratic when clustering sets containing up to 100,000s \order{s}.
In this setting, \textproc{RecAgglo} is faster than most categorical clustering algorithms.

\noindent\textbf{Non-optimal solution}: \textproc{RecAgglo} is non-deterministic and it does not produce a globally optimal clustering. This is due to the stochastic nature of the sampling process used \textproc{SampleClust}.
We show in Sect.~\ref{sec:perf_comparison} that clusters that we obtain during different runs are consistent. Also, their goodness is close to the one of clusters generated using plain agglomerative clustering, while improving on the basic sampling method for clustering.

\noindent\textbf{Hybrid clustering (using numerical features):}
Numerical features can be input to a clustering algorithm for continuous data (e.g., K-means, DBScan, etc.) in order to generate clusters in a standalone manner. The resulting clustering (cluster indexes) can be used as an additional categorical attribute that is input to \ourname in a cluster aggregation fashion~\cite{gionis2007clustering}.

%


%% file: weight.tex
\section{Attribute weighting strategies}
\label{sec:weight}

\textit{Hamming} distance and \textit{Jaccard Index} are the most widely used metrics for computing the distance between two elements $u$ and $v$ represented using categorical attributes~\cite{maimon2005data}.
We use \textit{Hamming} distance in our clustering algorithms since it is fast to compute. It counts the number of different attribute values between two elements:

\begin{equation}
\label{eq:hamming}
	Hamming(u,v) = \frac{1}{d} \sum_{i=1}^d w_i \times (u_i \neq v_i )
\end{equation}

By default, \textit{Hamming} like many other metrics gives the same weight to every attribute ($w_i = 1$).
However, different attributes might not contribute equally to quantifying the similarity between $u$ and $v$, or to produce ``good'' clusters.
For instance, if attributes having high cardinality are matching, this may indicate a higher similarity than if attributes having low cardinality are matching.
We propose two novel strategies for weighting attributes which capture these aspects and help  addressing \ref{chal:cardinality}.
The first strategy is based on feature cardinality while the second uses labels from known frauds and legitimate \orders.

\subsection{Cardinality driven attribute weight}
\label{subsec:cardinality_weight}

We define a function to compute the weight $w_i$ for an attribute $a_i$ based on its cardinality.
The cardinality $card_i$ is the total number of values attribute $a_i$ can take.
The rationale for weighting attributes based on cardinality is the following: the probability of two elements $u$ and $v$ having equal value for an attribute $i$ is inversely proportional to the attribute cardinality $card_i$ for uniformly distributed attribute values.
The goal of this weighting strategy is to give larger weights to attributes having high cardinality.

Cardinality of the attributes in our dataset is not bounded since values may be added as new \orders are made, e.g., new customers signing up.
Thus, we use the inverse normalized richness index~\cite{jost2006entropy} as the basis for weight computation: $R_i^{-1}=\frac{n_i}{card_i}$.
$n_i$ is the number of instances in a given set of size $N$ for which $a_i$ is not null .
$R_i^{-1}$ is a positive decreasing function of the attribute cardinality.

We use a sigmoid function ($\frac{x}{|1+x|}$) to scale $R_i^{-1}$ to $\mathit{[0,1]}$. We normalize its value over this range using the median value of $R_i^{-1}$ computed over all 37 attributes: $median(R^{-1})$.
Finally, we scale our weight to an intended range that controls the maximum difference between attribute weights.
We chose the range $\mathrange{1, 3}$ - we do not discard any attributes and a given attribute can have at most 3 times higher weight than any other.
We compute cardinality driven weights as follows:

\begin{equation}
\label{eq:cardinality_weight}
w^\#_i = 1 + 2 \times \left( 1 - \dfrac{R_i^{-1}} {median(R_i^{-1}) + R_i^{-1}} \right), w^\#_i  \in \mathrange{1,3}
\end{equation}


\subsection{Label driven attribute weight}
\label{subsec:label_weight}

We define the second function to compute weights using ground truth fraud labels of past \orders.
These weights are computed in order to satisfy two requirements for our method: \ref{goal:fraud} maximizing clustered fraud and \ref{goal:impurity} minimizing cluster impurity.

We start by clustering a set of \orders using default attribute weights ($w_i = 1$) using \textit{Hamming} distance.
We obtain clusters of three types: (a) pure clusters $c_f$ containing only frauds, (b) pure clusters $c_l$ containing only legitimate \orders and (c) mixed cluster $c_m$.
$c_m$ clusters violate \ref{goal:impurity} and their number must be minimized.
$c_f$ clusters contribute to \ref{goal:fraud} and their number must be maximized.

We aim to emphasize the importance of attributes that help generating $c_f$ clusters and de-emphasize the importance of attributes that do not by scaling their weight accordingly.
The higher the weight, the more important the attribute.
We compute the contribution of an attribute $a_i$ towards generating a cluster $c$ using the Simpson index~\cite{simpson1949measurement}.
It is defined as $\lambda_i(c) = \sum_{j=1}^{card_i} p_j^2$, where $p_j$ is the probability of encountering the attribute value $v_j$ in $c$: the ratio of elements having $v_j$ for $a_i$.
High Simpson index indicates that a low number of different values $v$ is present in the cluster. This means $a_i$ has significantly contributed towards generating this cluster.

Using the Simpson index we define two metrics $Adv_{f/l}$ in  Eq.~(\ref{eq:advfl}) and $Adv_{p/m}(a_i)$ in Eq.~(\ref{eq:advpm}).
$Adv_{f/l}(a_i)$ measures the \textit{advantage} of the attribute $a_i$ in generating pure fraudulent clusters $c_f$ rather than pure legitimate clusters $c_l$.
$Adv_{p/m}(a_i)$ quantifies the \textit{advantage} of the attribute $a_i$ in generating pure clusters $c_f$ and $c_l$ instead of mixed clusters $c_m$.
High $Adv_{f/l}$ helps achieving \ref{goal:fraud} and a high $Adv_{p/m}$ helps achieving \ref{goal:impurity}.
The normalization term $norm_{adv}(a_i)$ ensures that $Adv_{f/l}(a_i) + Adv_{p/m}(a_i) \in \mathrange{0,2}$ which allows us to keep the final weight in the range $\mathrange{1, 3}$

\begin{equation}
\label{eq:advfl}
Adv_{f/l}(a_i) = \dfrac{\mean{\lambda_i(c_f)} - \mean{\lambda_i(c_l)}}{norm_{adv}(a_i)}
\end{equation}

\begin{equation}
\label{eq:advpm}
Adv_{p/m}(a_i) = \dfrac{\mean{\lambda_i(c_f)} + \mean{\lambda_i(c_l)} - 2 \times \mean{\lambda_i(c_m)}}{2 \times norm_{adv}(a_i)}
\end{equation}

We compute our label driven weights using both these advantages as follows:

\begin{equation}
\label{eq:weight_label}
w^\ast_i = 1 + Adv_{f/l}(a_i) + Adv_{p/m}(a_i), w^\ast_i  \in \mathrange{1, 3}
\end{equation}


%% file: exp_setup.tex
\section{Performance metrics and datasets}
\label{sec:metric_datasets}

We discussed that \ourname meets~\ref{goal:cl_size} by design.
We empirically evaluate the remaining requirements~\ref{goal:impurity} (minimize cluster impurity), \ref{goal:fraud} (maximize clustered fraud) and \ref{goal:time} (minimize execution time).

\subsection{Performance metrics}


We evaluate~\ref{goal:impurity} by computing the cluster \textit{impurity} measure $I$, which is used to evaluate the quality of a clustering~\cite{gionis2007clustering}. We give the label of the majority class to each cluster and all samples that do not belong to this class are counted as the impurity.
For a clustering containing $k$ clusters of sizes $s_1,\ldots, s_k$ and the sizes of the majority class in each clusters $m_1,\ldots, m_k$, the impurity index is defined as:

\begin{equation}
I = \dfrac{\sum_{i=1}^k (s_i - m_i)}{\sum_{i=1}^k s_i} = \dfrac{\sum_{i=1}^k (s_i - m_i)}{n}
\end{equation}


We evaluate~\ref{goal:fraud} by calculating the clustered fraud rate ($CFR$) which is the ratio of clustered frauds to the total number of frauds.
For the count of frauds in each cluster $f_1,\ldots, f_k$ and the total number of frauds $F$, $CFR$ is defined as:

\begin{equation}
CFR = \dfrac{\sum_{i=1}^k (f_i)}{F}
\end{equation}

We evaluate~\ref{goal:time} by measuring the computation time $t$ of the clustering for the given dataset.

Our objectives are to minimize the impurity $I$ and computation time $t$ while maximizing the $CFR$.

\subsection{Datasets}

We use several datasets composed of real fraud and legitimate \orders placed on the Zalando website in 2017 and 2018.
Zalando receives on average 29 million orders per quarter~\cite{zalando2019}.
Our ground truth fraud labels are obtained based on actual payment status of the order 12 weeks after it is placed.
Orders without a label are considered legitimate.

The datasets presented in the following are sampled from the original \order data.
They differ in size and ratio of legitimate to fraudulent orders. We use them for different experiments that we describe as follows.



\noindent\textbf{Small datasets with artificial distribution.}
We sample two small datasets \frtrain and \frtest that are used for selecting hyperparamters of agglomerative clustering (Sect.~\ref{sec:weighting_eval}) and for comparing the performance of several categorical clustering techniques (Sect.~\ref{sec:perf_comparison}) respectively.
These sets are small enough for most categorical clustering techniques to run in a reasonable amount of time (\textless 10 hours).
Also, they contain enough frauds to generate many fraudulent clusters that we can use to compute sensible impurity $I$ and $CFR$ metrics.
Frauds are artificially over-sampled (1 fraud / 2 legitimate) compared to a real-world distribution.
Each dataset consists of 10 disjoint subsets, each composed of 10,000 legitimate \order{s} and 5,000 frauds.

\noindent\textbf{Large datasets with artificial distribution.}
We sample two larger datasets \gesmall and \gelarge that are used for selecting hyperparamters of \ourname (Sect.~\ref{sec:hyper_para_rec}). These also have an artificial distribution where frauds are over-sampled compared to the real-world distribution.
The imbalance is larger and more realistic in these datasets though (1 fraud / 5 legitimate  and 1 fraud / 19 legitimate).
\gesmall consists of 10 disjoint subsets, each composed of 25,000 legitimate \order{s} and 5,000 frauds.
\gelarge consists of 5 disjoint subsets, each composed of 95,000 legitimate and 5,000 fraud.
The composition of datasets with artificial distribution is presented in detail in App.~\ref{app:dataset}.

\begin{figure*}
    \centering
    \begin{tabularx}{\linewidth}{XXX}

			\includegraphics[width=0.69\columnwidth]{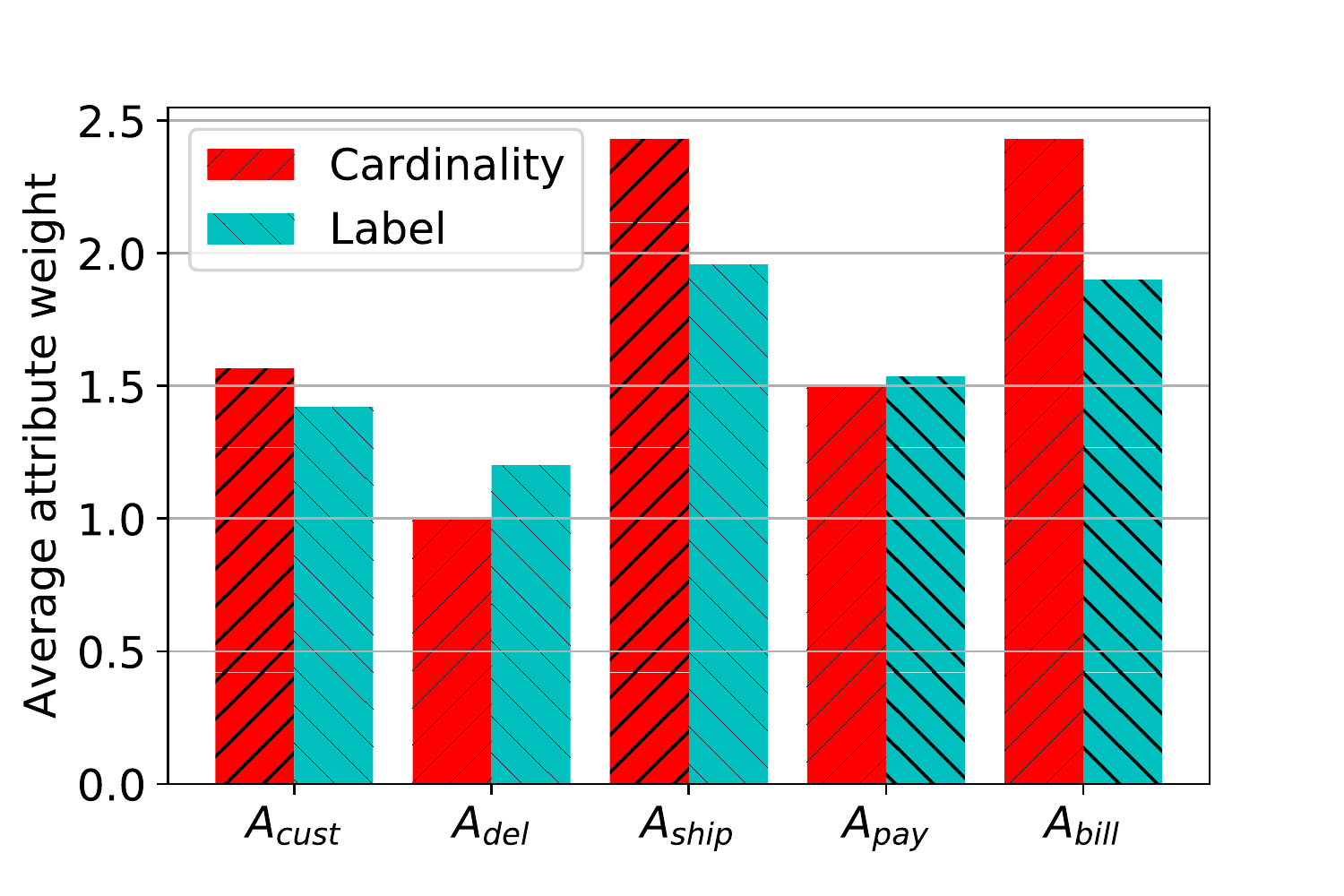}
                \caption{Average weights of five attribute categories using cardinality and label driven weights. Shipping and billing attributes are the most important in generating clusters.}
                \label{fig:weight_cat}
		&
			\includegraphics[width=0.69\columnwidth]{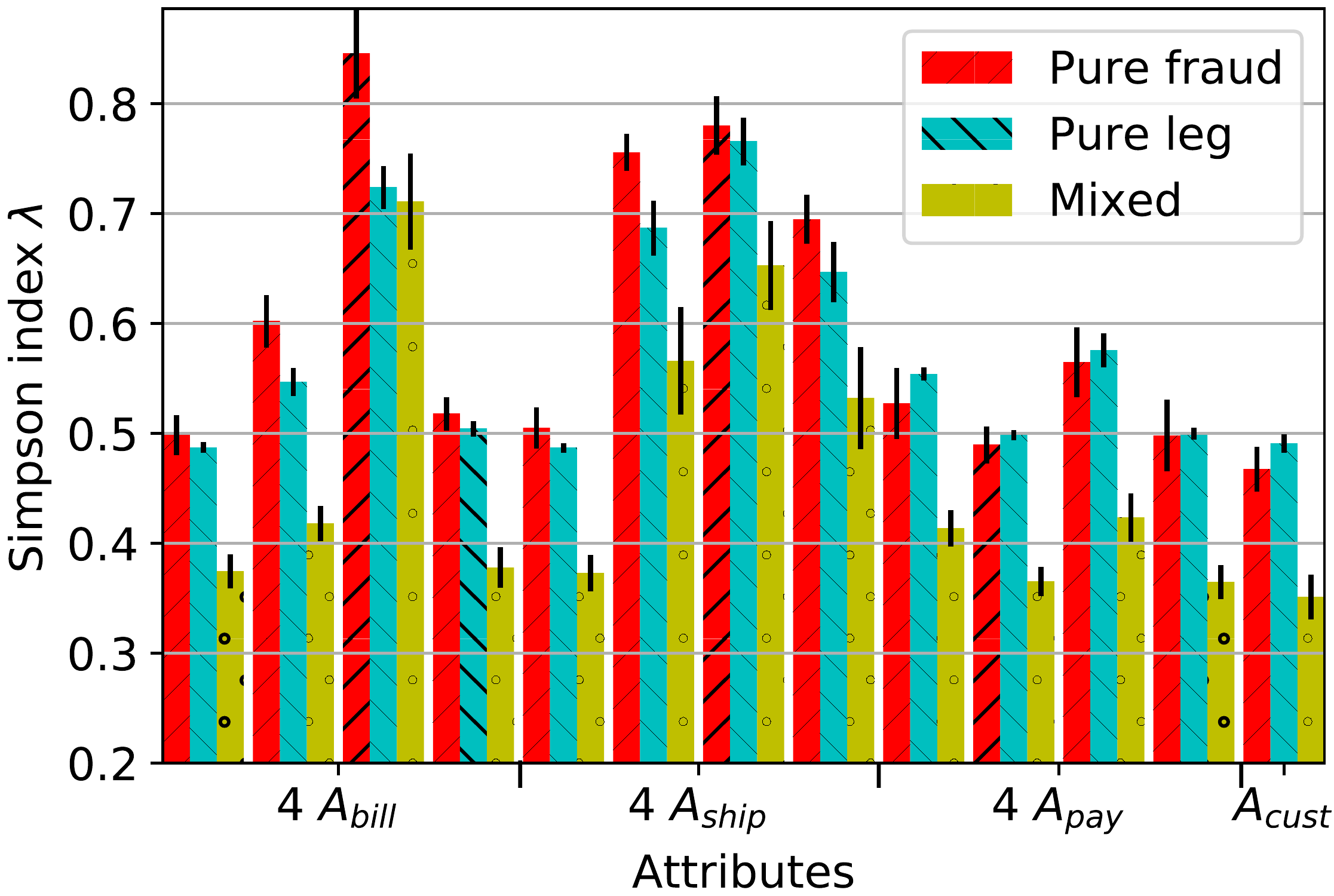}
                \caption{Simpson index $\lambda$ for 13 attributes providing the best advantage in generating fraudulent and pure clusters. Billing, shipping and payment attributes provide the best advantage.}
                \label{fig:simpson}
        &
                 \includegraphics[width=0.67\columnwidth]{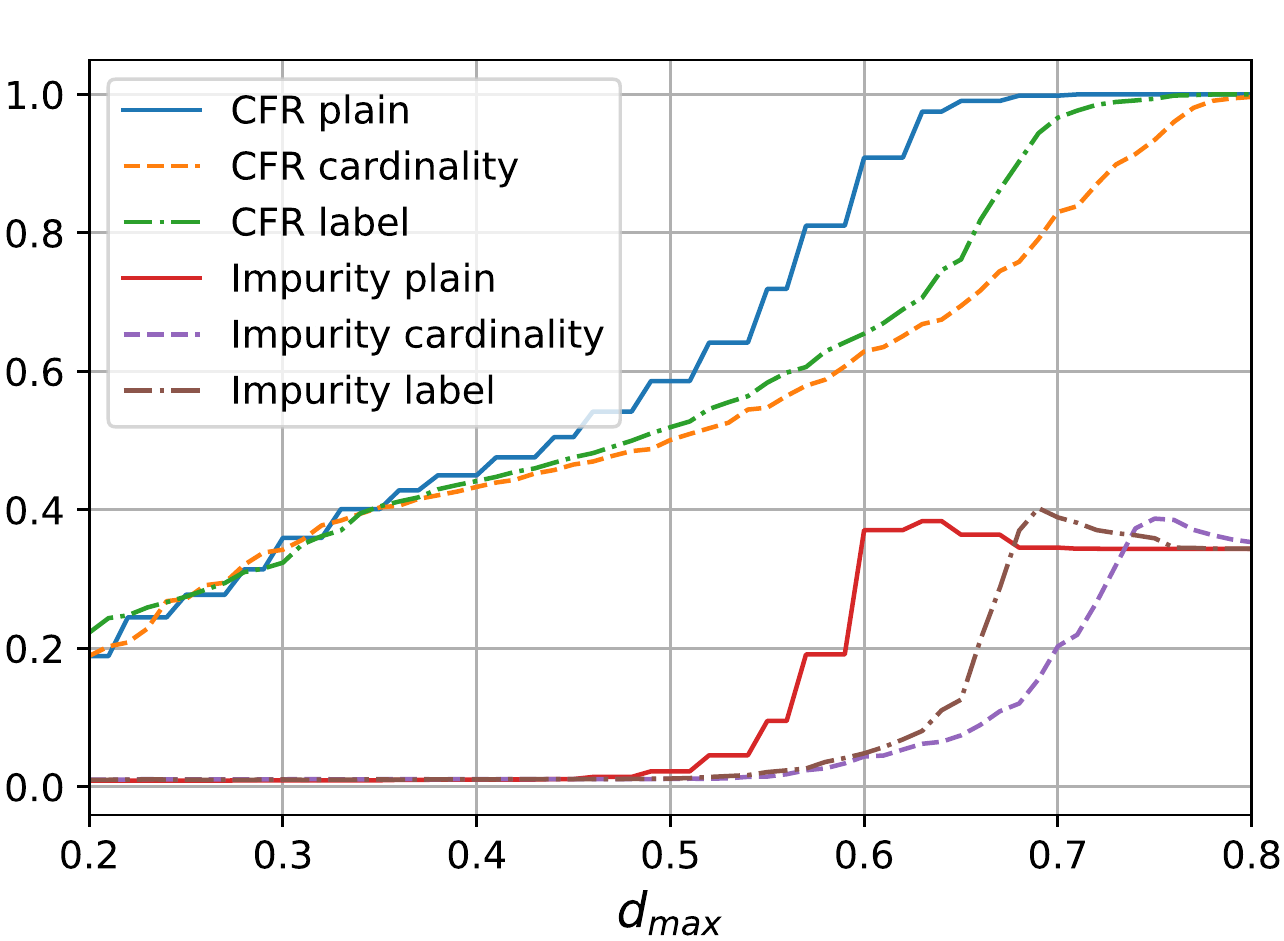}
         		\caption{Increase in Impurity and \textit{CFR} with $d_{max}$ for three attribute weighting strategies. A fine-grained choice of Impurity/CFR tradeoff is possible using cardinality and label driven weights.}
         \label{fig:weight_dmax}
    \end{tabularx}
\end{figure*}

\noindent\textbf{Real-world datasets.}
Finally, we select real-world datasets that will be used to evaluate the actual effectiveness of \ourname at clustering fraud in Sect.~\ref{sec:real-world-eval}.
These datasets consist of all Zalando Fashion Store orders placed between April 1st and May 5th 2018 (35 days) in Germany (\gereal), Switzerland (\hrreal), the Netherlands (\nlreal), Belgium (\bereal) and France (\frreal).
These datasets contain more than 6 million orders in total, with a realistic fraud/legitimate order ratio (well below 1\% before fraud cancellation).


Each of these datasets is complemented with a background dataset containing only frauds placed between January 1st and March 31st 2018 (90 days). These datasets are respectively named \gebg, \hrbg, \nlbg, \frbg \bebg.

%% file: exp_agglo.tex
\section{Weighting strategies evaluation}
\label{sec:weighting_eval}

Agglomerative clustering is the basis for \ourname. It uses three hyperparameters: a distance metric, a linkage method and the maximum distance for cluster fusion $d_{max}$.
Recall that we selected \textit{Hamming} as a distance metric because of its low computation cost.
We selected the \textit{single} linkage method based on evaluation described in App.~\ref{subapp:linkage_eval}.
We want to select the optimal weighting strategy and a distance $d_{max}$ which minimize the impurity $I$ and maximize the $CFR$.
We compare the default weighting strategy ($w_i = 1$) to the cardinality $w^\#_i$ and label driven weights $w^\ast_i$ we introduced in Sect.~\ref{sec:weight}.


%

\subsection{Weight computation}
\label{sec:weight_compute}

We select a random sample of 2M \orders placed in France in 2017 to compute our cardinality driven weights using Eq.~(\ref{eq:cardinality_weight}).
In this subset, we obtain a minimum inverse normalized richness index $min(R_i^{-1}) = 1.606$ for one of the attributes in the $A_{cust}$ category. It means that the same value repeats less than twice (on average) for over 2M samples. On the other hand, we obtained $max(R_i^{-1}) = 1,000,000$ for one of the attributes in $A_{del}$ meaning it has only two possible values.
These statistics highlight the imbalance in the cardinality of attributes representing \orders (\ref{chal:cardinality}), which justifies cardinality based weighting strategy.
We computed the value for $median(R_i^{-1}) = 149$ that we use to calculate the weights of all 37 attributes.

\begin{center}
$w\#_i = 1 + 2 \times (1 - \dfrac{R_i^{-1}} {149 + R_i^{-1}})$
\end{center}


We start by clustering each set \frtrain-i using agglomerative clustering and default weights $w_i = 1$ to compute our label driven weights. We select the maximum distance for cluster fusion $d_{max} = 0.56$, which generates a clustering with impurity $I = 0.095$ and $CFR = 0.719$ (App.~\ref{subapp:linkage_eval}). The majority of fraud is clustered (72\%) and there is a significant number of mixed clusters as depicted by the high impurity (9.5\%).
We compute the Simpson index $\lambda_i$ for each attribute $a_i$ in each generated cluster. We aggregate these results to compute the mean $\mean{\lambda_i}$ for pure fraudulent clusters ($c_f$), pure legitimate clusters ($c_f$) and mixed clusters ($c_m$).
Using these statistics we compute our advantage metrics (Eq.~(\ref{eq:advfl}) and (\ref{eq:advpm})) and by extension our final label driven weights.

\subsection{Attribute importance}
\label{sec:weight_analysis}

Figure~\ref{fig:weight_cat} depicts the average cardinality and label driven weights of attributes in each category: $A_{cust}$, $A_{del}$, $A_{pay}$, $A_{ship}$ and $A_{bill}$.
Despite the different rationale and implementation for our two weighting strategies, we see they give similar high and low weights to the same attributes.
Attributes in $A_{ship}$ and $A_{bill}$  have the largest weights according to both strategies.
These attributes differ between customers ($A_{ship}$) and between \orders ($A_{bill}$), which explains their high cardinality and their large cardinality driven weights.


Figure~\ref{fig:simpson} depicts the averaged Simpson index for the 13 attributes providing the highest \textit{advantage}.
We observe that $A_{bill}$ attributes give the best advantage for generating fraudulent rather than legitimate clusters (higher Simpson index in pure fraudulent clusters). $A_{ship}$ attributes also provide a small advantage towards that goal, while $A_{pay}$ and $A_{cust}$ do not.
Different values of the Simpson index depict the advantage of each attribute and are captured by our label driven features. It can be seen that $A_{ship}$ and $A_{bill}$ attributes have the highest weights (Fig.~\ref{fig:weight_cat}).
On the other hand, all 13 attributes contribute to generating pure clusters (lower Simpson index for mixed clusters).
$A_{cust}$ and $A_{del}$ contribute the least to our \textit{advantages} and they have the lowest weight in Fig.~\ref{fig:weight_cat}.

High Simpson index values for $A_{bill}$ and $A_{ship}$ attributes indicate that fraudulent \orders have more similar billing and shipping information than legitimate \orders.
On the other hand, there is no significant difference for $A_{cust}$ and $A_{pay}$ attributes.
These results might indicate that fraudsters tend to use several user accounts and payment methods with similar billing and shipping information.
Consequently, our generated clusters have characteristics that are typically associated with fraud campaigns as presented in Sect.~\ref{sec:organized_fraud}.


\subsection{Weighting strategies performance}
\label{sec:weight_perf}


We clustered the 10 \frtrain-i datasets using default attribute weights, cardinality and label driven weights.
Each weighting strategy provides a similar $I$/$CFR$ tradeoff that is detailed in App.~\ref{subapp:weighting_perf}.
Nevertheless, label driven weights provide a slightly better $CFR$ than other strategies for the same impurity value and we select it for the remaining experiments.
A more interesting property of cardinality and label driven weights can be observed in Fig.~\ref{fig:weight_dmax}. Both these strategies offer a smoother increase of impurity and \textit{CFR} while varying $d_{max}$. In contrast, default weights have abrupt changes and long plateaus providing the same performance. In this setting it is difficult to select an optimal $d_{max}$ that provides desired impurity and $CFR$ values.
Cardinality and label driven weights can be used to effectively fine-tune $d_{max}$ in order to achieve desired performance characteristics.
Using Fig.~\ref{fig:weight_dmax}, we select $d_{max} = 0.5$ with label driven weights, which results in an average impurity $I = 0.012$ for a $CFR = 0.52$.
With this value, impurity remains low (about 1\%) while more than half of the frauds are clustered.

\textit{Hamming} distance with label driven weights, single linkage and distance $d_{max} = 0.5$ are used in \ourname in all following experiments.



%
%


%% file: exp_recursive.tex
\section{\ourname performance evaluation}
\label{sec:eval_recursive}

We evaluate the performance of \ourname in terms of impurity, $CFR$ and computation time when clustering real online \order{s}.
We compare this performance to several state-of-the-art categorical clustering techniques to show \ourname is best suited for this task.

\subsection{Hyperparameter setting}
\label{sec:hyper_para_rec}

\ourname requires defining $\delta_a$ and \textproc{SampleClust} requires defining $\rho_s$ and $\rho_{mc}$ (cf. Sect.~\ref{sec:design}).
We compute optimal hyperparameter values with the primary goal of minimizing computation time and the secondary goals of minimizing impurity $I$ and maximizing $CFR$.
We set $\delta_a = 1,000$ with computation time being the only consideration in mind.
Agglomerative clustering takes 25s to process 1,000 samples. Consequently, this is an upper bound for the computation time of \textproc{AggloClust} in \ourname.
The upper bound for the fall back agglomerative clustering is given for $4,000$ ($4 \times \delta_a$) elements to cluster and takes 388s.

We perform a grid search over $\rho_s=\lbrace 0.25,0.5,1,2 \rbrace$ and $\rho_{mc}=\lbrace 1.01,1.5,2,3,4,6,10 \rbrace$ to select hyperparameter values for \textproc{SampleClust}.
We run \ourname with every hyperparameter combination on \gesmall (10 runs) and \gelarge (5 runs) computing $I$, $CFR$ and computation time $t$.
A detailed analysis of the grid search results on \gesmall is presented in App.\ref{subapp:sampleClust-para}. It shows that a too low $\rho_{mc}$ value (e.g., $1.01$) or a high $\rho_s$ value significantly increases the computation time of \ourname.
We selected $\rho_s = 0.5$ and $\rho_{mc} = 6$ as these hyperparamters provide the best tradeoff with $t=4,110s$, $CFR=34.0\%$ and $I=3.0\%$ on \gelarge.
The sample size for sampling is $0.5 \times \sqrt{n}$ and the maximum number of clusters to generate is \textit{maxclust} $= n/6$.


\subsection{Experimental setup}
\label{sec:exp_setup_recagglo}

We use four categorical clustering algorithms to compare the performance of \ourname: \textproc{AggloClust}, \textproc{SampleClust}, \textproc{Kmodes} and ROCK.
We ran experiments on a consumer grade laptop with 8GB of RAM and Intel Core i5 (2.7GHz) processor.
We already presented \textproc{AggloClust} and \textproc{SampleClust} in Sect.~\ref{sec:agglo_sampling}.

\textproc{Kmodes}~\cite{huang1998extensions} is an extension of the \textit{Kmeans} algorithm for categorical data.
It starts by selecting $k$ random points as starting ``modes'' and calculates the distance between each element-mode pair.
It assigns each element to the cluster that has the closest pair-wise distance to its mode.
Modes and clusters are updated  over several iterations.
\textproc{Kmodes} uses \textit{Hamming} as distance metric and its time complexity is $O(n \times k)$. We use the PyPi implementation of \textproc{Kmodes}~\cite{kmodes:python}.

\textit{ROCK}~\cite{guha2000rock} is a clustering algorithm that uses a concept of ``neighbor''.
Two elements are neighbors if the distance between them is lower than a threshold $\theta$.
Then, if two elements have enough common neighbors they are placed in the same cluster.
ROCK uses the \textit{Jaccard index} as distance metric and its time complexity is $O(n^2 \times log(n)+ n^2 )$.
We adjust the ROCK implementation from~\cite{Novikov2019} to accommodate for categorical data.

\subsection{Performance analysis}
\label{sec:perf_comparison}

\begin{table}
\caption{Impurity, $CFR$, and computation time for 4 categorical clustering algorithms. Results are averaged over 10 runs on \frtest (15,000 samples). *: results for \textproc{ROCK} are computed on 5,000 randomly picked samples. \ourname generates the clusters with the lowest impurity in a short time.}
\label{tab:algo_comp}
\begin{tabular}{lccc} \hline
Algorithm & $I$ (\%) & $CFR$ (\%) & time \\ \hline
\textproc{RecAgglo}$_{\delta_{max}=0.5}$ & \textcolor{green}{0.8} & 42.1 & \textcolor{green}{185s} \\ \hline
\textproc{AggloClust}$_{\delta_{max}=0.5}$ & 1.2 & \textcolor{green}{51.9} & \textcolor{red}{1h31} \\ \hline
\textproc{SampleClust}$_{\delta_{max}=0.5,\rho_s=0.5}$ & 1.1 & \textcolor{red}{1.2} & 38s \\
\textproc{SampleClust}$_{\delta_{max}=0.6,\rho_s=0.5}$ & 3.3 & \textcolor{red}{2.2} & 38s \\
\textproc{SampleClust}$_{\delta_{max}=0.6,\rho_s=2}$ & 2.4 & \textcolor{red}{7.7} & 158s \\ \hline
\textproc{Kmodes}$_{k=1,000}$ & \textcolor{red}{20.5} & 99.8 & 20m \\
\textproc{Kmodes}$_{k=5,000}$ & \textcolor{red}{14.1} & 91.6 & 1h31 \\
\textproc{Kmodes}$_{k=12,000}$ & \textcolor{red}{10.5} & 39.2 &  7h44\\ \hline \hline
*\textproc{ROCK}$_{\theta=0.55,t=0.45}$ & 7.1 & 51.4 &  \textcolor{red}{3h08}\\
*\textproc{ROCK}$_{\theta=0.45,t=0.40}$ & \textcolor{green}{0.9} & 30.3 &  \textcolor{red}{1h48} \\ \hline

\end{tabular}
\end{table}

We cluster the 10 \frtest subsets (15,000 \order{s} each) and average $I$, $CFR$ and $t$ over the 10 runs.
\textproc{ROCK} was not able to cluster 15,000 elements in a reasonable amount of time  and its results are computed on a random sample of 5,000 elements from \frtest.
These results are summarized in Tab.~\ref{tab:algo_comp}
\textproc{Kmodes} is not able to generate a clustering with low impurity ($I > 0.1$) despite the high number of clusters we try to generate (up to $k=$12,000 as generated by \textproc{AggloClust}).
\textproc{SampleClust} generates a clustering with low impurity in short time. However, it clusters only a very small number of frauds ($CFR < 0.08$).
\textproc{ROCK} produces a clustering with low impurity $I = 0.009$ and higher $CFR = 0.303$ but its computation time on 5,000 samples is prohibitive (\textgreater 1h30).
\textproc{AggloClust} is the algorithm providing the best trade-off between impurity and $CFR$. It clusters over half of the frauds with an impurity close to 1\%.
Its computation time of 1h31 is prohibitive for a fairly small dataset.

\ourname has the second lowest computation time (3 minutes) and the clusters it generates have the lowest impurity of all algorithms ($I = 0.008$). \ourname achieves high $CFR = 0.421$, which is only 10 percentage points lower than the $CFR$ of \textproc{AggloClust}.
\ourname meets \ref{goal:impurity} with its low impurity and \ref{goal:fraud} with its relatively high $CFR$.

\begin{figure}[th]
                \centering
                \includegraphics[width=0.95\columnwidth]{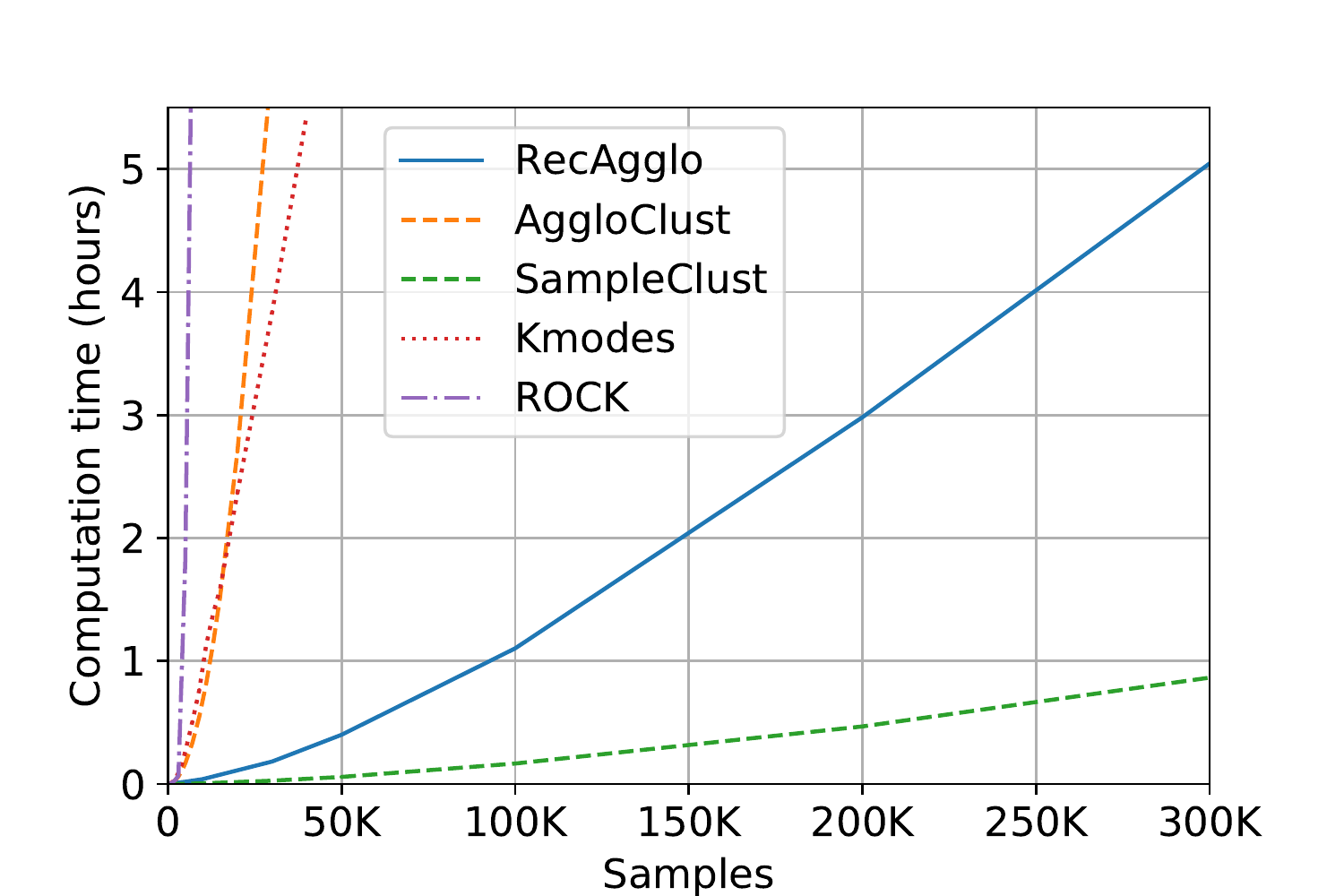}
                \caption{Computation time (averaged over 5 runs) vs. sample size for 5 clustering algorithms. Only \textproc{SampleClust} and \ourname can scale to large datasets.}
                \label{fig:timing}
\end{figure}

We timed the clustering of an increasing set of \orders from \gereal (up to $n=300,000$) to assess the scalability of these algorithms. We report the running time averaged over 5 runs in Fig.~\ref{fig:timing}.
We use settings from Tab.~\ref{tab:algo_comp} providing the lowest computation time: \textproc{SampleClust} (${\delta_{max}=0.5,\rho_s=0.5}$)  and \textproc{ROCK} (${\theta=0.45,t=0.40}$).
For \textproc{Kmodes}, we use ${k=min(n/2, 5000)}$, bounding $k$ to 5,000 to limit the complexity of \textproc{Kmodes} (function of $k$).

\textproc{ROCK}, \textproc{AggloClust} and \textproc{Kmodes} cannot process 50,000 \order{s} in less than 5 hours. Their complexity is at least quadratic, which prevent scaling to large datasets.
\textproc{SampleClust} is the fastest algorithm, it clusters 300,000 samples in less than one hour.
Finally, we see that \textproc{RecAgglo} scales to medium-size datasets despite its high worst-case complexity (cf. Sect.~\ref{sec:algo_properties}). In our specific use case of clustering \order{s}, \ourname is much faster than algorithms with quadratic complexity because it requires only 2-5 recursions (instead of $n$ in the worst case) to obtain a final clustering that meets our goodness criterion.
It is able to cluster 100,000 \order{s} in one hour and its computation time increases almost linearly with the set size: one hour more per additional 50,000 \order{s}.
\ourname processes 300,000 samples in 5 hours, which is the number of \order{s} received by Zalando daily. \ourname addresses \ref{chal:scale} and it meets the low computation time requirement \ref{goal:time}: it can be used in a real-world deployment setting to cluster frauds.

%% file: exp_real_world.tex
\section{Real-world fraud detection}
\label{sec:real-world-eval}

We assess the performance of \ourname for our applied cases of preventing organized fraud: \ref{app:screening} prioritizing screening and \ref{app:automated} automated fraud cancellation.
We take the following real-world scenario and we evaluate it using our real-world datasets.
We take a set of unlabeled \orders $O_u$ placed over one day (e.g., in a 24h window).
We have access to a set of labeled fraudulent \orders $O_f$ from an earlier time period (e.g., more than one day old).
Our goal for \ref{app:screening} is to provide human operators with clusters containing a maximum number of frauds from $O_u$.
Our goal for \ref{app:automated} is to use the resulting clusters to automatically and reliably detect a maximum number of organized frauds.
This empirical evaluation assesses the performance of \ourname in its own right. It assumes a deployment in parallel of any existing detection system (as depicted in Fig.~\ref{fig:pipeline}) and does not take into account the fraud detection pipeline that Zalando has already developed.
The evaluation of the incremental value of \ourname with respect to the existing Zalando pipeline is not in the scope of this paper.


\subsection{Prioritizing screening}
\label{sec:eval_A1}

To effectively prioritize screening we must provide human analysts with a manageable amount of orders that contains a large portion of frauds.
If we screen only clusters generated by \ourname, these clusters must contain a large amount of fraud (maximize $CFR$) and a low amount of legitimate \order{s} (minimize $CLR$).
$CLR$ is the ratio of clustered legitimate \order{s}, equivalent to $CFR$ but for legitimate \order{s}.
Considering our real-world scenario, we cluster \order{s} in $O_u \cup O_f$ using \ourname.
We cluster the merged sets together since frauds from $O_u$ can be related to frauds from $O_f$.
We want our clusters to maximize $CFR_{u}$, which is the $CFR$ computed solely on $O_u$, i.e., the portion of fraudulent \orders form $O_u$ that are clustered.
Recall that we want our clusters to have low impurity $I$ and to keep a low $CLR$ to reduce the workload of human analysts.

%


\begin{table}
\caption{Performance of \ourname at clustering real-world \orders from 5 countries. Different delays for obtaining fraud labels (1 day/30 days). Cluster impurity ($I$), ratio of unlabeled ($CFR_{u}$) and overall ($CFR$) frauds clustered. Ratio of legitimate \order{s} clustered ($CLR$).  Frauds are clustered 3-7 times more than legitimate \orders. Clusters do not mix legitimate \order{s} and frauds.}
\label{tab:realTime}

\begin{tabular}{lcccc|cccc} \hline
	& \multicolumn{4}{c}{1 day delay for labels} & \multicolumn{4}{c}{30 days delay for labels} \\
Dataset & $I$ & $CFR$ & $CFR_{u}$ & $CLR$ & $I$ & $CFR$ & $CFR_{u}$ & $CLR$\\ \hline
\gereal &  0.9 & 51.9 & 43.4 & 9.6 & 0.8 & 52.8 & 33.3  & 9.6 \\
\hrreal &  2.9 & 49.8 & 45.8 & 17.7 & 2.6 & 52.1 & 37.3  & 17.8\\
\nlreal &  1.2 & 34.8 & 34.8 & 7.9 & 1.0 & 33.1 & 26.0  & 7.9\\
\bereal & 1.2 & 49.8 & 41.8 & 13.0 & 1.1 & 48.6 & 29.9  & 13.0 \\
\frreal &  0.3 & 50.4 & 44.1 & 7.1 & 0.2 & 47.7 & 32.1  & 7.1 \\ \hline
Overall & 1.3 & 49.7 & 43.5 & 10.9 & 1.1 & 50.2 & 33.7 & 11.0 \\ \hline
\end{tabular}

\end{table}

We take \orders from each day in the datasets \gereal, \hrreal, \nlreal, \bereal and \frreal to create 35 sets $O_u$ per country.
We create associated sets $O_f$ of frauds from a prior period of 60 consecutive days using \gebg, \hrbg, \nlbg, \frbg and \bebg.
We consider two scenarios for composing $O_f$ that depict the delay in obtaining fraud labels. The first scenario assumes a one day delay meaning that $O_f$ contains frauds from the period of 60 days ending when $O_u$ starts. The second scenario assumes a 30 days delay meaning that $O_f$ contains frauds from a period ending 30 days before $O_u$ ends. We selected 30 days because it is a sufficient delay for preliminary identification of \orders being in payment default.
We cluster the 35 resulting sets $O_u \cup O_f$ for each country and present averaged $I$, $CFR$, $CFR_u$ and $CLR$ per country in Tab.~\ref{tab:realTime}.

\ourname produces clusters having low impurity which varies depending on the country. Notably, there is a 10-fold difference between Switzerland ($I = 2.9\%$) and France ($I = 0.3\%$).
This shows that legitimate \orders from, e.g., Switzerland or Belgium, are more similar to frauds than this is the case in France or Germany.
The $CLR$ is 3- to 7-times lower than the $CFR$ showing that \ourname clusters more frauds than legitimate \orders.
The $CFR_u$ is around 35-45\%, which is slightly lower than the $CFR$ but consistently larger than the $CLR$ (7-18\%).
For screening prioritization \ref{app:screening}, around 90\% of legitimate \orders could be discarded from manual analysis while preserving around 40\% of frauds that could be detected.
\ourname can be used to prioritize screening.

We see that the increased delay in obtaining fraud labels decreases the $CFR_u$ by around 10 percentage points.
There is a strong time dependence between \orders belonging to the same cluster and to the same fraud campaign. Obtaining fraud labels in a timely manner is crucial to maximize our ability to detect organized fraud.
It is worth noting that while we ran \ourname once per day in this experiment, it can be re-ran continuously as new \orders are received (using a sliding window containing one day of \orders $O_u$). We recall that \ourname processes 100,000 \orders in one hour.


\subsection{Automated fraud cancellation}

We devise a simple technique to automatically detect and cancel frauds using clusters generated by \ourname.
We detect an unlabeled \order from $O_u$ as fraud (1) if it is clustered and (2) if at least one known fraud from $O_f$ belongs to this cluster.
We call this technique \textit{label propagation} - known fraud propagates its label to the whole cluster it belongs to.

We apply this technique to the clusters generated in Sect.~\ref{sec:eval_A1}.
For each country in Tab.~\ref{tab:detection}, we report $Recall$, $Precision$ and False Positive Rate ($FPR$) averaged over 35 clustering results (35 days).
$Recall$ is the ratio of correctly detected frauds ($TP$) over the total number of frauds ($TP + FN$) and it represents the ability to detect frauds.
$Recall_{clust}$ is computed only on clustered frauds, while $Recall_{final}$ is computed on all frauds.
$Recall_{final} = Recall_{clust} \times CFR_u$ is the actual rate of detected frauds.
$Precision$ is the ratio of correctly detected frauds ($TP$) over the total number of detected frauds ($TP + FP$), i.e., the reliability of fraud detection.
$FPR$ is the ratio of legitimate \orders incorrectly detected as fraud ($FP$) over the total number of legitimate \orders ($TN + FP$). It corresponds to the error rate for legitimate \orders.

\begin{table}
\caption{Recall, precision and false positive rate (FPR) for automated fraud detection in 5 countries. One quarter of frauds are detected while generating a few false alarms (0.1\%). Only 35.3\% of detected frauds are actual frauds.}
\label{tab:detection}
\begin{tabular}{lccccc} \hline
Dataset &  $Recall_{clust}$ & $Recall_{final}$ & $Precision$ & $FPR$ \\ \hline
\gereal &  59.8 & 26.2 & 35.9 & 0.1 \\
\hrreal &  72.3 & 33.2 & 17.0 & 0.3 \\
\nlreal &  60.4 & 20.5 & 29.3 & 0.1 \\
\bereal & 63.6 & 26.5 & 34.4 & 0.2 \\
\frreal &  65.8 & 30.0 & 71.4 & 0.1 \\  \hline
Overall & $62.6 (\pm 10.6)$ & $26.4 (\pm 5.5)$ & $35.3 (\pm 6.3)$ & $0.1 (\pm 0.03)$ \\  \hline

\end{tabular}
\end{table}


This simple automated detection technique would cancel over one quarter ($Recall_{final} = 26.4\%$) of fraud, which likely represents 62.6\% ($Recall_{clust}$) of all organized fraud.
It also generates few false alarms ($FPR = 0.1\%$) for legitimate \orders.
Despite this very low $FPR=0.1\%$, the precision remains low because of the large imbalance between fraud and legitimate \order{s} ($\#$\textit{fraud} $\ll \# legitimate$). Our average precision of 35.4\% means that 2/3 \order{s} detected as fraud are actually legitimate \order{s}.
This low precision is prohibitive for automated fraud cancellation~\ref{app:automated} but it can prioritize screening~\ref{app:screening}.
Human analysts could cancel over 25\% of frauds with little effort required.

We investigated further the characteristics of legitimate \order{s} that our \textit{label propagation} technique incorrectly identified as \fraud.
We computed the ratio of these \order{s} that belong to four legitimate categories namely, (1) \textit{fully} and (2) \textit{partially} \textit{returned} to the retailer (where a customer does not pay for all items and returns some of them), (3) \textit{partly unpaid} (where items in the \order remain unpaid while delivered) and (4) \textit{canceled} by the customer.
We observed that 94.7\% of the false positives that degrade the Precision of \textit{label propagation} belong to one of these four categories. 
The majority of the false positives (63.9\%) are returned \orders while 24.3\% are partly unpaid \orders.


\subsection{Evading fraud detection}
\label{sec:evasion}

Recent research in adversarial machine learning~\cite{papernot2018sok,juuti2019prada} has shown that machine learning-based systems can be evaded by manipulating their inputs~\cite{goodfellow2014explaining}.
Fraudsters can evade our fraud detection approach by making their \order{s} less similar to one another. This would result in \ourname not being able to group together \fraud{s} from the same campaign.
\ourname  quantifies the similarity between \order{s} by computing the Hamming distance.
If two attributes have different values, the distance between two \order{s} increases and their similarity decreases.
Our attributes have a string representation. The inequality between two attribute values can be obtained by modifying a single character from one of them.
Such a modification (e.g., typo in street address) may have no impact on successfully placing and receiving an \order (fraud purpose), while its similarity to other \order{s} could be greatly reduced.
An adversary can modify \order attributes that are input to \ourname to evade fraud detection.

This limitation can be addressed by using only attributes that are resilient to adversarial manipulations in \ourname~\cite{marchal2017off}. Resilient attributes are those for which manipulations inherently defeat the fraud purpose. For instance, a small modification to a credit card number makes payment information inconsistent, which causes rejection of payment and of the \order. Credit card number is an attribute resilient to adversarial manipulations.
Alternatively, we can use a metric more fine-grained than binary equality for string comparison. The edit distance can accurately quantify the similarity between two strings and it accommodates small adversarial modifications.
An adversary would have to make large modifications to reduce the similarity between two orders, which makes evasion harder.
Nevertheless, the edit distance is expensive to compute and it will increase the running time of \ourname clustering.

%% file: related-work.tex
\section{Related Work}\label{sec:related-work}

\subsection{Categorical clustering}

Categorical clustering faces two main challenges: scalability and guarantee of convergence.
Generic algorithms such as \textproc{Kmodes}~\cite{huang1998extensions}, \textit{ROCK}~\cite{guha2000rock}, \textit{CACTUS}~\cite{ganti1999cactus} (greedy hierarchical grouping of tuples) or \textit{CLICKS}~\cite{zaki2005clicks} (representing the dataset as a graph and finding disjoint vertices) provide good guarantees of convergence and many recent clustering algorithms~\cite{gionis2007clustering, bendechache2016efficientls} are built upon these techniques.
The main alternative approaches use information-theoretic criteria to assess the quality of the clustering.
For instance, \textit{COOLCAT}~\cite{daniel2002coolcat} searches locally to find clusters with the lowest entropy, while \textit{LIMBO}~\cite{andritsos2004limbo} produces a hierarchical summary of the data that preserves as much information as possible.
While generalizable to any kind of categorical data, these algorithms have high complexity and they do not scale to large datasets.


To improve scalability, the data can be either partitioned into chunks that are clustered individually and then combined into global clusters~\cite{bendechache2016efficientls} or it can be transformed into representations that make processing faster, e.g., Merkle trees~\cite{liu2007treesclustering}. These approaches are typically problem specific (images~\cite{liu2007treesclustering}, streamed data~\cite{guha2003clusteringstreams}) and they are not applicable to categorical data.
A more generic solution produces several clusterings using subsets of features and aggregates the results to obtain global clusters~\cite{gionis2007clustering}. Similarly, low-dimensional clusters can be generated using dissimilarity matrices and then combined using an ensemble method to get the final clustering~\cite{amiri2018ensembling}. These methods use sampling, as we do, but they take a local approach, trying to reduce the dimension of the input space to improve scalability.

\subsection{Fraud detection}

Fraud detection is a sparse subject relevant to many domains: credit card fraud~\cite{gomez2018neuralnetsfraud, carcillo2017spark}, tax evasion~\cite{Bogdanov2015estonia}, online dating~\cite{tangil2019dating}, erotic content~\cite{hutchings2019understanding}, advertising~\cite{nagaraja2019advertising}, among others. Despite efforts to systemize it~\cite{sorin2012fraudsurvey, bolton2002fraudreview, niu2019comparison}, there is no commonly accepted means of comparing different detection techniques or application scenarios.

Many solutions try to detect frauds in isolation using supervised classifiers such as neural networks~\cite{gomez2018neuralnetsfraud,aleskervo1997cardwatch, Brause1999neuraldatamining, maes2002bayesandneural} and ensemble of decision trees~\cite{carcillo2017spark}.
Improvements have been proposed to incorporate the time component in the process of eCommerce fraud detection by identifying changes in the underlying distribution of \order{s} (also known as \textit{concept drift})~\cite{mao2018adaptive}.
Finally, one can analyze the temporal activity of users from a small set of features to predict possible account take-over~\cite{halawa2018suspicious}.
These solutions analyze \order{s} in isolation or in a group related to a specific user and they are not suitable for identifying organized fraud that involves many users. Also, they require labeled data that is not available for fraud campaigns.

Graph-based methods can be used to take a global view at the fraud detection problem~\cite{tang2010socialnetwork, molloy2017graph, akoglu2009anomalydetection, sangers2018pagerank}.
Social network analysis methods~\cite{tang2010socialnetwork, akoglu2009anomalydetection}, the PageRank algorithm~\cite{sangers2018pagerank} and sets of graph-derived features~\cite{molloy2017graph} have been used to spot frauds in the payment network and to identify the key links between frauds.
However, these methods require having access to (or constructing) a graph with a well-defined notion of the vertex and edge (e.g. credit card and merchant vertices and interaction as an edge). In our case, we do not have a proper link between \order{s} that we could use to build a graph. Instead, we try to identify similarities through clustering. Moreover, building a graph with a static set of assumptions could hinder the performance of the graph analysis method if the nature of fraud changes overtime.

A few works also proposed to use clustering to identify fraudulent activity, as we do.
However, these methods either use only numerical features as input~\cite{thiprungsri2011clusteranalysis} or they do not scale to data of such a scale as ours (100,000s samples)~\cite{lesot2012hybrid}.


%% file: conclusion.tex
\section{Conclusion}

We introduced a novel clustering solution (\ourname) to detect organized fraud and we evaluated it on 6M real-world orders placed on the Zalando website.
We showed \ourname is able to process 100,000s of orders in a few hours and it groups over 40\% of fraudulent \order{s} together. The algorithm can be deployed and used to efficiently prioritize screening \ref{app:screening}.
We further proposed a simple technique named \textit{label propagation} that uses our generated cluster to automatically detect 26.2\% of fraud while raising false alarms for only 0.1\% of legitimate \orders.
In spite of its high accuracy, \textit{label propagation} incorrectly identifies many legitimate \orders as \fraud (35\% precision). Considering our definition of what ``fraud'' is, \textit{label propagation} cannot be used for automated fraud cancellation \ref{app:automated}.
Nevertheless, we observed that 95\% of the legitimate \orders incorrectly identified as \fraud by \textit{label propagation} are either returned, partly unpaid or canceled \orders.

Canceling legitimate \orders from good customers can create a lot of harm to a business~\cite{false-decilne:2017}.
It generates a bad customer experience with a detrimental effect on customer satisfaction and customer lifetime value, which may ultimately decrease the profitability of an online retail service.
``\Fraud'' has a subjective definition that is different for different online retailers.
The effectiveness and deployability of a fraud detection system are evaluated according to this definition.
Hence, the suitability of our solution to prioritize screening \ref{app:screening} and automatically cancel fraud  \ref{app:automated} depends on the categorization of returned (fully/partly), unpaid (fully/partly) and canceled \orders for a given online retailer.

%

%% file: app-clustering.tex
\section{Clustering algorithms details}
\label{app:clustering}

\subsection{Agglomerative clustering}
\label{app:agglo_clust}

The agglomerative clustering algorithm is detailed in Algorithm~\ref{algo:aggloclust}.
\textproc{DistanceMatrix} compute a $|c_i| \times |c_j|$ matrix of the distance between elements of $c_i$ and $c_j$.
\textit{Hamming} distance and \textit{Jaccard Index} are the most widely used~\cite{maimon2005data} methods for measuring the distance of categorical data. We select \textit{Hamming} distance since it is fast to compute. It counts the number of different attribute values between two elements $u$ and $v$ (cf. Eq.~\ref{eq:hamming}).

The linkage matrix computed using \textproc{LinkageMatrix} describes the successive cluster fusions to go from singletons to a single cluster (dendrogram).
Cluster fusion is done according to the distance between two clusters, which is defined by a \textit{linkage} method. \textit{Single linkage} uses the minimum distance between any two points in each cluster.
\textit{Complete linkage} takes the maximum distance between any two points from each cluster. \textit{Ward linkage}~\cite{maimon2005data} takes the increase in the sum of square obtained by merging two clusters rather than by keeping them separate.
We use the single linkage method here.
We compute the final clustering using \textproc{Cluster} based on the linkage matrix and the \textit{'distance'} criterion parametrized by $d_{max}$.

\begin{algorithm}[ht]
\caption{Agglomerative clustering}
\label{algo:aggloclust}
\begin{algorithmic}[1]

	\State{
		Let $c = {v_1, \ldots, v_m}$ denote a set of $|c| = m$ elements $v$, \\
		$d_{max}$, the maximum distance for cluster fusion.
		} \\

	\Function{AggloClust}{$c, d_{max}$}

		\State{$D \gets$ \textproc{DistanceMatrix}($c, c$, 'Hamming')}
		\State{$LM \gets$ \textproc{LinkageMatrix}($D$, 'single')}
		\State{$C_{res} \gets$ \textproc{Cluster}($LM, d_{max}$, 'distance')}

		\State{return $C_{res}$}

	\EndFunction
\end{algorithmic}
\end{algorithm}

\subsection{Agglomerative clustering with sampling}
\label{app:sample_clust}

The algorithm for agglomerative clustering with sampling is presented in Algorithm~\ref{algo:sampling}.
We modify the basic sampling algorithm to create a maximum number of clusters.
Rather than initially clustering sampled elements, we use each of them as the basis for a new cluster (no initial clustering).
Our sampling algorithm randomly selects $n$ samples from $c$ (where $n$ is a factor $\rho_s$ of $\sqrt{|c|}$). Thus, we compute the distance matrix between the sample and all elements in $c$.
We use the single linkage method to compute the linkage matrix since we only know the distance to a single element in each cluster (because of sampling).
Finally, it generates a clustering $C_{res}$ using the maximum number of clusters criterion (\textit{'maxclust'}) where $c_{max}$ is computed as a factor $\rho_{mc}$ of the set size $|c|$.
This criterion is selected with the goal of splitting a large set of elements into smaller clusters but without providing any guarantee of goodness for the resulting clustering.

\begin{algorithm}[ht]
\caption{Agglomerative clustering with sampling}
\label{algo:sampling}
\begin{algorithmic}[1]

	\State{
		Let $c = {v_1, \ldots, v_m}$ denote a set of $|c| = m$ elements $v$, \\
		$\rho_s$, the multiplying factor for the sample size,\\
		$\rho_{mc}$, the dividing factor for maxclust number.
		} \\

	\Function{SampleClust}{$c, \rho_s, \rho_{mc}$}
		\State{$n \gets \rho_s \times \sqrt{|c|}$}
		\State{$c_{max} \gets m \div \rho_{mc}$}

		\State{$s \gets$ \textproc{RandomSample}($c, n$)}
		\State{$D \gets$ \textproc{DistanceMatrix}($s, c$, 'Hamming')}
		\State{$LM \gets$ \textproc{LinkageMatrix}($D$, 'single')}
		\State{$C_{res} \gets$ \textproc{Cluster}($LM, c_{max}$, 'maxclust')}

		\State{return $C_{res}$}
	\EndFunction

\end{algorithmic}
\end{algorithm}

%% file: app-dataset.tex
\section{Datasets composition}
\label{app:dataset}

\noindent\textbf{Small datasets with artificial distribution:}
We selected \frtrain and \frtest from Zalando \order{s} passed in France over 2017.
In each dataset, we simulate an artificial distribution where \fraud{s} are over-sampled compared to a real-world distribution (2 legitimate / 1 fraud).
\frtrain consists of 10 disjoint subsets \frtrain-i, each composed of 10,000 legitimate and 5,000 \fraud{s}.
The 5,000 \fraud{s} are randomly selected from a continuous period of 1-1.5 month.
The 10,000 legitimate \order{s} are randomly selected from a period of 1-2 days.
For each subset $i$, the period from which legitimate \order{s} are selected is included into the period from which \fraud{s} are selected.
\frtest consists of 10 disjoint subsets \frtest-i selected the same way as for \frtrain (10,000 legitimate and 5,000 \fraud{s}) \frtest and \frtrain are disjoint.

\noindent\textbf{Large datasets with artificial distribution.}
We select \gesmall and \gelarge from \order{s} passed in Germany over 2017.
Each subset of \gesmall is composed of 25,000 legitimate and 5,000 \fraud{s}.
The 5,000 \fraud{s} are randomly selected from a continuous period of 1 month.
The 25,000 legitimate \order{s} are randomly selected from a period of 1 day contained in the month from which \fraud{s} are selected.
Each subset of \gelarge is composed of 95,000 legitimate and 5,000 \fraud{s}.
The 5,000 \fraud{s} are randomly selected from a continuous period of 1 month.
The 95,000 legitimate \order{s} are randomly selected from a period of 1 week  contained in the month from which \fraud{s} are selected.

%% file: app-weighting.tex
\section{Hyperparameter selection}
\label{app:weighting}


\subsection{Linkage method selection}
\label{subapp:linkage_eval}

\begin{figure}[th]
                \centering
                \includegraphics[width=0.90\columnwidth]{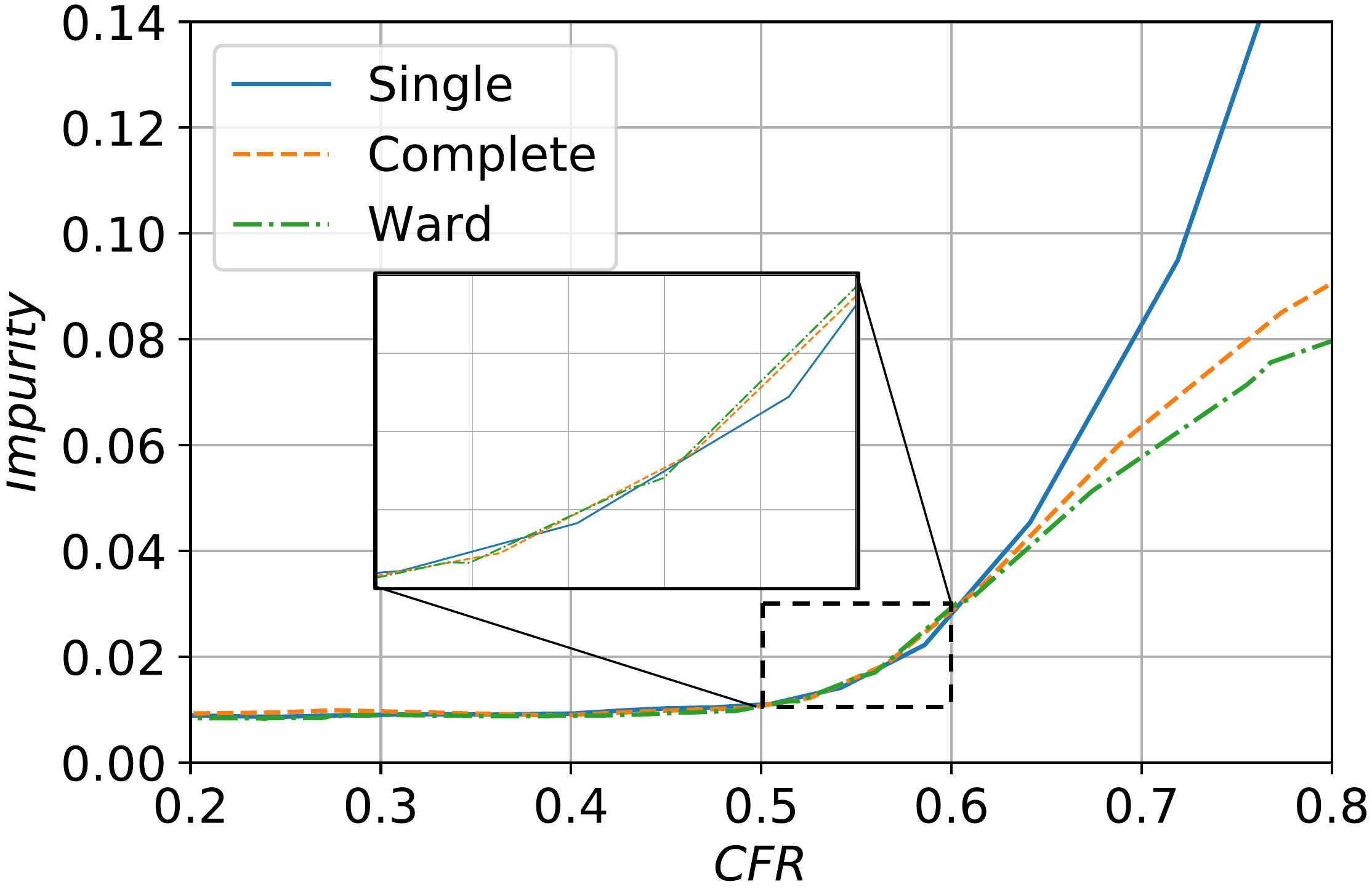}
         			\caption{Impurity vs. \textit{CFR} for single / complete / Ward linkage methods. Each linkage method provides the same Impurity/\textit{CFR} tradeoff when the impurity is low: 0.01-0.03.}
                \label{fig:linkage}
\end{figure}

We want to select a linkage method that minimizes the impurity $I$ and maximizes the clustered fraud rate ($CFR$).
We cluster the 10 \frtrain-i datasets using \textit{single} linkage, \textit{complete} linkage and \textit{Ward} linkage.
Figure~\ref{fig:linkage} shows the evolution of impurity $I$ according to $CFR$ while varying the maximum distance for cluster fusion $d_{max}$.
Values are averaged over 10 clustering results.
We see that for $CFR < 0.6$ all linkage methods have similar impurity values.
For $CFR > 0.6$ \textit{Ward} outperforms other linkage methods, while \textit{single} becomes the worst method with high increase in impurity.

Our primary goal is to keep the impurity as low as possible.
All methods are comparable at providing a high $CFR$ while keeping the impurity low (0.01 - 0.03).
\textit{Single} and \textit{complete} linkage are computed using a single distance between two points from two clusters: the closest and the furthest away ones respectively. Thus, they are faster to compute than \textit{Ward}. They are also better suited for clustering with sampling, since the distance between any two points is not available using sampling.
We can see from Fig.~\ref{fig:linkage} (zoom) that single linkage provides higher $CFR$ than complete linkage for the same impurity value.
Thus, we select \textit{single} linkage as our base linkage technique.

\begin{figure}[th]
                \centering
			\includegraphics[width=0.90\columnwidth]{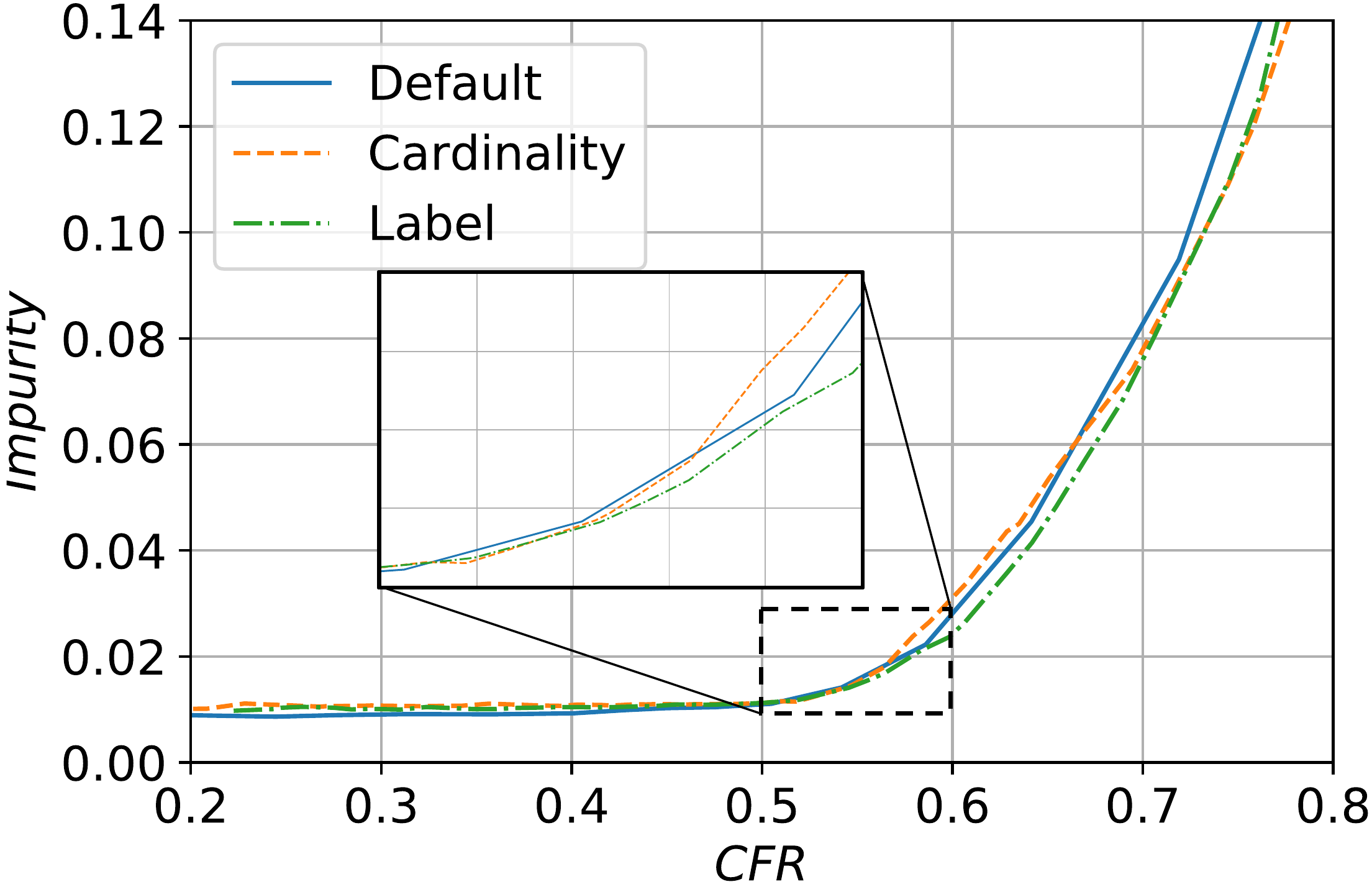}
         \caption{Impurity vs. \textit{CFR} for default / cardinality / label driven attribute weighting. All methods have comparable performance. Label driven weighting provides marginally higher \textit{CFR} for the same impurity.}
         \label{fig:weight_CFR}
\end{figure}

\begin{table}
\caption{Results of grid search for \textproc{SampleClust} hyperparameters: $\rho_s=\lbrace 0.25,0.5,1,2 \rbrace$ and $\rho_{mc}=\lbrace 1.01,1.5,2,3,4,6,10 \rbrace$. Average impurity, $CFR$ and computation time computed over 10 runs of \ourname on \gesmall ($\delta_a = 1,000$). A low performance score means a combination of hyperparameters that maximizes our clustering objectives: low impurity, high CFR and low computation time.}
\label{tab:rho_value}
\begin{tabular}{rrrrrr} \hline
$\rho_{mc}$ & $\rho_s$ & $I$ (\%) & $CFR$ (\%) & time (s) & performance score \\ \hline
\multirow{4}{*}{1.01} & 0.25 & 3.61 & 28.59 & 2,070 & 3.06 \\
 & 0.5 & 3.65 & 29.54 & 4,722 & 5.10 \\
 & 1 & 3.75 & 30.13 & 9,049 & 9.33 \\
 & 2 & 3.68 & 30.53 & 16,657 & 15.90 \\ \hline
\multirow{4}{*}{1.5} & 0.25 & 3.43 & 27.72 & 461 & 1.17 \\
 & 0.5 & 3.45 & 28.36 & 657 & 1.00 \\
 & 1 & 3.38 & 28.49 & 1,010 & \textcolor{green}{0.89} \\
 & 2 & 3.49 & 28.74 & 1,365 & 1.61 \\ \hline
\multirow{4}{*}{2} & 0.25 & 3.38 & 27.45 & 397 & 0.98 \\
 & 0.5 & 3.39 & 28.29 & 620 & \textcolor{green}{0.72} \\
 & 1 & 3.43 & 28.48 & 920 & 1.09 \\
 & 2 & 3.53 & 28.70 & 1,300 & 1.82 \\ \hline
\multirow{4}{*}{3} & 0.25 & 3.43 & 27.42 & 410 & 1.30 \\
 & 0.5 & 3.42 & 28.31 & 629 & \textcolor{green}{0.86} \\
 & 1 & 3.44 & 28.82 & 913 & 0.91 \\
 & 2 & 3.56 & 28.75 & 1,262 & 1.91 \\ \hline
\multirow{4}{*}{4} & 0.25 & 3.46 & 27.67 & 417 & 1.29 \\
 & 0.5 & 3.51 & 28.65 & 632 & 1.12 \\
 & 1 & 3.40 & 28.71 & 884 & \textcolor{green}{0.72} \\
 & 2 & 3.48 & 28.76 & 1,265 & 1.49 \\ \hline
\multirow{4}{*}{6} & 0.25 & 3.49 & 27.56 & 456 & 1.57 \\
 & 0.5 & 3.44 & 28.48 & 682 & \textcolor{green}{0.88} \\
 & 1 & 3.43 & 28.76 & 936 & \textcolor{green}{0.88} \\
 & 2 & 3.48 & 28.78 & 1,350 & 1.52 \\ \hline
\multirow{4}{*}{10} & 0.25 & 3.41 & 27.89 & 540 & 0.99 \\
 & 0.5 & 3.44 & 28.43 & 720 & 0.94 \\
 & 1 & 3.47 & 28.81 & 995 & 1.12 \\
 & 2 & 3.53 & 28.89 & 1,456 & 1.83 \\ \hline
\end{tabular}
\end{table}

\subsection{Weighting strategy selection}
\label{subapp:weighting_perf}

Figure~\ref{fig:weight_CFR} shows the evolution of impurity $I$ according to $CFR$ while varying the maximum distance for cluster fusion $d_{max}$.
Values are averaged over 10 clustering results.
We see that for $CFR < 0.55$ all weighting strategies have similar impurity values.
For $CFR > 0.55$ our label driven weighting becomes better than other strategies providing up to a 2 percentage points higher $CFR$ than other strategies while keeping impurity low (0.025).
Cardinality driven features provide the best trade-off between $CFR$ and impurity (for $I > 0.10$). This is an interesting result but it is not useful since our goal is to minimize impurity.
We see that label driven weighting gives the best trade-off between $CFR$ and impurity.

%
%

%% file: app-samplClust.tex
\subsection{\textproc{SampleClust} hyperparameters selection}
\label{subapp:sampleClust-para}

Table~\ref{tab:rho_value} presents the impurity ($I$), $CFR$ and computation time ($t$) for each combination of \textproc{SampleClust} hyperparameters $\rho_{mc}$ and $\rho_s$ tested during the grid search. These results are limited to the grid search we performed on \gesmall and they depict how our performance metrics vary according to $\rho_{mc}$ and $\rho_s$ values.
We also computed a \textit{performance score} to choose the optimal hyperparameter combination. It is the sum of normalized performance metrics $\hat{I}$, $\hat{CFR}$ and $\hat{t}$, which are respectively defined in Eq.~(\ref{eq:norm-I}), (\ref{eq:norm-CFR}) and (\ref{eq:norm-t})\footnote{We discarded time results obtained using $rho_{mc}$ from the normalization process in Eq.~(\ref{eq:norm-t}). These are too high and represent outliers. $max(t)$ and $min(t)$ only take $\rho_{mc}=\lbrace 1.5,2,3,4,6,10 \rbrace$ into account.}.
A low performance score depicts a combination of $\rho_{mc}$ and $\rho_s$ that maximizes our clustering objectives: low impurity, high CFR and low computation time.

\begin{equation}
\label{eq:norm-I}
\hat{I}_{(\rho_s=x,\rho_{mc}=y)} = \dfrac{I_{(\rho_s=x,\rho_{mc}=y)} - min(I)}{max(I) - min(I)}
\end{equation}

\begin{equation}
\label{eq:norm-CFR}
\hat{CFR}_{(\rho_s=x,\rho_{mc}=y)} = \dfrac{max(CFR) - CFR_{(\rho_s=x,\rho_{mc}=y)}}{max(CFR) - min(CFR)}
\end{equation}

\begin{equation}
\label{eq:norm-t}
\hat{t}_{(\rho_s=x,\rho_{mc}=y)} = \dfrac{t_{(\rho_s=x,\rho_{mc}=y)} - min(t)}{max(t) - min(t)}
\end{equation}

We see that hyperparamater choice heavily impacts the computation time while impurity and $CFR$ remain almost constant. Too low $\rho_{mc}$ value (e.g., $1.01$) or a high $\rho_s$ value significantly increase the computation time. This is expected since $\rho_s$ defines the sample size used in clustering with sampling. A high $\rho_s$ value means a large sample.